\documentclass[10pt,twocolumn,letterpaper]{article}

\usepackage[pagenumbers]{cvpr} 
\usepackage{booktabs}
\usepackage{graphicx}
\usepackage{multirow}

\usepackage{amsmath,amssymb,bm}
\usepackage{float}
\usepackage{colortbl}
\usepackage{pifont}
\newcommand{\cmark}{\ding{51}}
\newcommand{\xmark}{\ding{55}}
\definecolor{oursrow}{RGB}{222,235,247}


\definecolor{cvprblue}{rgb}{0.21,0.49,0.74}
\usepackage[pagebackref,breaklinks,colorlinks,allcolors=cvprblue]{hyperref}

\def\confName{CVPR}
\def\confYear{2027}

\title{Looped-WAM: Efficient Robotic Manipulation via Looped World Action Modeling}

\author{Jingpeng Yang*, Yikun Wang*, Panjun Liu, Xinyao Zhuang, Weiye Xu, \\
Jiajun Deng$^\dagger$, Wengang Zhou$^\dagger$, Houqiang Li\\
{\small University of Science and Technology of China}\\
{\small $\ast$: Equal contribution \,\, $\dagger$: Corresponding authors}\\
{\small yang.jp@mail.ustc.edu.cn}\\
{\small \textbf{Project page:} \url{https://looped-wam.github.io/}}%
}

\begin{document}
\maketitle


\begin{abstract}
World Action Models (WAMs) have attracted increasing attention for their ability to jointly model robot actions and their visual consequences. However, their substantial computational and memory demands hinder efficient deployment.
Existing acceleration strategies primarily focus on step distillation, feature reuse, and token pruning, with less attention to the parameter efficiency of the underlying architecture. 
In this paper, we propose \textbf{Looped-WAM}, a parameter-efficient WAM built on a looped Transformer architecture that shares block parameters across recurrent iterations.
Looped-WAM matches the performance of larger models with fewer parameters and a smaller memory footprint, while allowing test-time computation to be adjusted through recurrent depth. 
We further observe that recurrent refinement reduces the number of robot execution steps required for task completion. We introduce an adaptive computation mechanism that varies recurrent depth during inference to further reduce execution steps while maintaining task success rates.
Extensive evaluations on the LIBERO-Plus and RoboTwin 2.0 simulation benchmarks, together with real-world manipulation experiments, demonstrate the effectiveness of Looped-WAM in balancing efficiency and manipulation performance.
These results establish Looped-WAM as a new baseline for compact and compute-adaptive WAMs, providing a foundation for further exploration of recurrent architectures for robotic manipulation.

\end{abstract}
\section{Introduction}
\label{sec:intro}

\begin{figure}[t]
    \centering
    \includegraphics[width=0.50\textwidth]{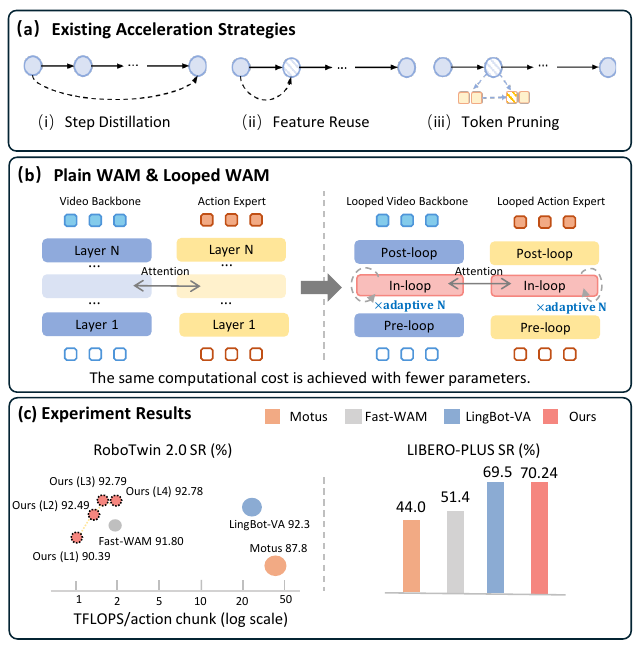}
    \vspace{-0.3cm}
    \caption{\textbf{Overview of Looped-WAM.} (a) Existing acceleration strategies use step distillation, feature reuse, or token pruning. (b) Looped-WAM replaces untied Transformer stacks in both the video backbone and action expert with a Pre-loop, shared In-loop, and Post-loop organization, decoupling parameter count from computational depth. (c) Success rate versus inference FLOPs on RoboTwin 2.0 (left) and mean success rate on LIBERO-Plus (right). L1--L4 denote fixed loop counts of one to four. The LIBERO-Plus bar labeled Ours corresponds to the fixed four-loop model (70.24\%).}
    \label{fig:overview}
    \vspace{-0.5cm}
\end{figure}

World Action Models (WAMs)~\cite{yuan2026fastwam,wang2026openwam,lv2026realtimewam} have recently emerged as a compelling paradigm for robotic manipulation by jointly modeling visual world dynamics and robotic actions. By pairing a pretrained video generation backbone~\cite{wan2025wan} with an Action Expert for action prediction, this joint modeling scheme has yielded promising performance across diverse manipulation tasks.

However, in standard Transformer-based WAMs, both the video backbone and the Action Expert consist of deeply stacked blocks with untied parameters, tightly coupling computational depth with model size. This static design not only incurs a substantial memory footprint, but also fixes the computational graph, preventing the model from flexibly scaling test-time compute according to state or task complexity. To mitigate runtime overhead, prior studies across vision, language, and generative models have explored inference-time acceleration strategies such as step distillation, token pruning~\cite{rao2021dynamicvit,chen2024fastv}, token merging~\cite{bolya2023tome}, feature caching~\cite{ma2024deepcache,zhao2026fasterwam}, and early-exit mechanisms~\cite{teerapittayanon2016branchynet,xin2020deebert,schwartz2020right}. Yet, these post-hoc or execution-level optimizations still operate within untied, parameter-heavy backbones: they leave intrinsic parameter redundancy untouched and cannot extend iterative refinement beyond a predefined physical depth.



To address these limitations, we propose \textbf{Looped-WAM} (Fig.~\ref{fig:overview}), a parameter-efficient World Action Model built upon weight-shared recurrent computation~\cite{dehghani2019universal,lan2020albert,yang2024looped,saunshi2025latent}. Our core insight is that repeatedly applying parameter-shared Transformer blocks enables progressive refinement of complex spatiotemporal dynamics and action representations without inflating parameter count. Specifically, Looped-WAM adopts a unified Pre-loop / In-loop / Post-loop architecture across both the video backbone and the Action Expert, where recurrent In-loop blocks iteratively refine joint latent representations. By decoupling effective computational depth from physical model size, Looped-WAM substantially reduces memory overhead while supporting flexible test-time compute budgets.

Furthermore, we observe that iteratively refining action representations through recurrent In-loop blocks yields more accurate control trajectories, thereby reducing the number of physical execution steps required for task completion. Motivated by this finding, we design an adaptive computation mechanism that combines deep supervision across loop iterations with a lightweight adaptive exit gate. This design dynamically allocates recurrent iterations conditioned on current state complexity, minimizing both per-step inference latency and overall execution steps without compromising task success rates.



We systematically evaluate Looped-WAM across diverse robotic manipulation tasks on RoboTwin~2.0~\cite{mu2025robotwin} and LIBERO-Plus~\cite{fei2026liberoplus}. Experimental results demonstrate that Looped-WAM matches or outperforms larger WAM baselines~\cite{yuan2026fastwam,li2026efficientwam,zhao2026fasterwam,li2026lightwam} with a substantially smaller memory footprint and higher inference efficiency. Moreover, our in-depth analysis of recurrent depth and adaptive computation validates that Looped-WAM achieves superior efficiency--performance trade-offs under varying compute budgets, establishing recurrent computation as an effective paradigm for compact and compute-adaptive World Action Models.

Our main contributions are summarized as follows:
\begin{itemize}
    \item We present \textbf{Looped-WAM}, a compact World Action Model built on a unified Pre-loop / In-loop / Post-loop architecture. Sharing Transformer weights across loop iterations in both the video backbone and the Action Expert decouples computational depth from model size, cutting memory footprints without sacrificing performance.

    \item We uncover that iterative latent refinement yields more precise actions that shorten physical task execution steps. Building on this finding, we pair loop-wise deep supervision with a lightweight exit gate to dynamically adjust recurrent depth at test time according to state complexity.

    \item Across RoboTwin~2.0~\cite{mu2025robotwin}, LIBERO-Plus~\cite{fei2026liberoplus}, and real-world physical manipulation tasks, Looped-WAM matches or outperforms larger WAM baselines at a fraction of their parameter cost, while reducing both per-step inference latency and overall execution steps.
\end{itemize}

\section{Related work}
\label{sec:related_work}

\subsection{World action models}

World Action Models (WAMs) integrate visual world dynamics modeling with robotic action prediction, leveraging pretrained video generation backbones~\cite{wan2025wan} to provide rich spatiotemporal representations for embodied control. This paradigm differs from Vision-Language-Action (VLA) models~\cite{brohan2023rt2,kim2024openvla,black2024pi0}, which typically map observations and language instructions directly to actions without explicit future-state modeling. Recent WAM frameworks, such as OpenWAM~\cite{wang2026openwam} and Cosmos Policy~\cite{kim2026cosmospolicy}, adapt large-scale video diffusion models for visuomotor planning and control, exploring diverse strategies for coupling visual prediction with action decoding. Despite their strong generalization and physical reasoning capabilities, inheriting billion-scale video backbones incurs substantial memory footprints and computational latency, hindering real-time robotic deployment.

\begin{figure*}[t]
    \centering
    \includegraphics[width=0.8\textwidth]{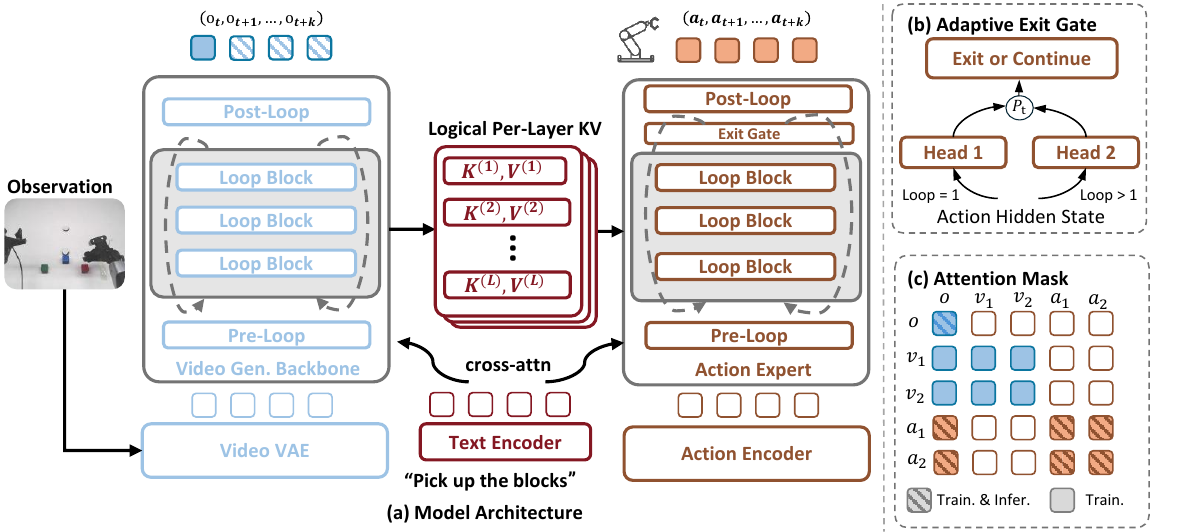}
    \caption{\textbf{Overview of Looped-WAM.} (a) The video backbone and action expert share a Pre-loop / In-loop / Post-loop organization and jointly model future video and action during training. At inference, one video prefill pass per observation, including a predefined number of recurrent iterations, produces logical layer-wise KV features that are cached and reused across action denoising steps. (b) The adaptive exit gate uses separate heads for the first and subsequent loop iterations. (c) The attention mask for observation ($o$), future-video ($v$), and action ($a$) tokens. Hatched entries are active during both training and inference, while solid-filled entries are active only during training.}
    \label{fig:architecture}
    \vspace{-0.3cm}
\end{figure*}

\subsection{Efficient world action modeling}

Recent efforts to improve WAM efficiency generally fall into two categories: inference acceleration and architectural compression. To reduce inference latency without altering backbone capacity, Fast-WAM~\cite{yuan2026fastwam} bypasses explicit future-video synthesis at test time, RealtimeWAM~\cite{lv2026realtimewam} combines consistency distillation~\cite{song2023consistency} with asynchronous execution, and Faster-WAM~\cite{zhao2026fasterwam} exploits sparse video--action interactions alongside intermediate feature reuse. While effective at lowering runtime cost, these methods retain the massive parameter counts of the underlying video models. Conversely, lightweight architectures reduce parameter overhead directly: Efficient-WAM~\cite{li2026efficientwam} pairs a compact video expert with asymmetric denoising, and Light-WAM~\cite{li2026lightwam} adopts a scaled-down video backbone with direct action decoding. However, shrinking network depth or width often compromises representational capacity for complex dynamics. More broadly, token pruning~\cite{rao2021dynamicvit,chen2024fastv}, token merging~\cite{bolya2023tome}, feature caching~\cite{ma2024deepcache}, and early exiting~\cite{teerapittayanon2016branchynet,xin2020deebert,schwartz2020right} have been widely adopted to trim redundant computation in Transformers and diffusion models. Rather than trading model depth for parameter compactness or solely accelerating fixed-size backbones, our approach couples cross-depth parameter sharing with adaptive early exiting to achieve both parameter efficiency and dynamic inference speedups.

\subsection{Parameter sharing and looped Transformers}

Cross-layer parameter sharing decouples computational depth from parameter count by reusing weights across sequential refinement steps. Universal Transformers~\cite{dehghani2019universal} and ALBERT~\cite{lan2020albert} pioneered recurrent depth-wise computation in language modeling, while recent studies on looped Transformers~\cite{yang2024looped,saunshi2025latent} demonstrate that iterative weight reuse strengthens algorithmic reasoning and latent refinement. In generative vision, Looped-DiT~\cite{chng2026loopeddit} brings recurrent parameter sharing to diffusion Transformers via a pre-loop/shared-loop/post-loop design stabilized by intermediate deep supervision, and LiFT~\cite{derakhshani2026lift} introduces continuous-depth regression targets to extrapolate beyond training horizons. Extending recurrent computation from unimodal generation to multimodal visuomotor control presents unique challenges due to asymmetric modality dynamics. To address this, we introduce looped parameter sharing jointly across the video backbone and action expert of WAMs, integrating coupled video--action deep supervision with an adaptive exit gate tailored for iterative action denoising.

\section{Method}
\label{sec:method}
In this section, we introduce \textbf{Looped-WAM}, a recurrent World Action Model~\cite{yuan2026fastwam,wang2026openwam} built upon a Mixture-of-Transformers~\cite{liang2024mixture} backbone for joint video and action modeling. By recurrently applying a parameter-shared Transformer block within each modality-specific branch~\cite{dehghani2019universal,yang2024looped}, Looped-WAM increases the effective computational depth without introducing additional layer-specific parameters, enabling progressive refinement of latent representations within a single forward pass.

\subsection{Architecture}
\label{sec:model_architecture}

As illustrated in Fig.~\ref{fig:architecture}, Looped-WAM
builds upon the dual-tower architecture of WAM, comprising
a video backbone and an action expert. The two branches
maintain modality-specific parameters while exchanging
information through joint self-attention. We preserve
this cross-modal interaction mechanism while reorganizing
each branch into three sequential stages, namely the
\emph{Pre-loop}, \emph{In-loop}, and \emph{Post-loop}.
This three-stage organization is applied
to both branches while retaining joint video-action attention.
The Pre-loop performs initial feature processing,
the In-loop iteratively refines the latent representations,
and the Post-loop produces the final predictions.

\textbf{Training.} Formally, during training, given noisy video latents $\mathbf{z}_{t_v}$,
noisy action tokens $\mathbf{a}_{t_a}$, visual observations
$\mathbf{o}$, and a language instruction $\ell$, the
forward computation is formulated as
\begin{equation}
\label{eq:looped_forward}
\begin{aligned}
\mathbf{h}^{(0)}
    &= \mathcal{P}_{\theta_p}
       \bigl(\mathbf{x}_{t};\mathbf{o},\ell,\mathbf{t}\bigr),\\
\mathbf{h}^{(k+1)}
    &= \mathcal{B}_{\theta_b}
       \bigl(\mathbf{h}^{(k)};\mathbf{o},\ell,\mathbf{t}\bigr),\\
\hat{\mathbf{v}}
    &= \mathcal{Q}_{\theta_q}
       \bigl(\mathbf{h}^{(N)}\bigr),
\end{aligned}
\end{equation}
where $\mathbf{x}_{t}=(\mathbf{z}_{t_v},
\mathbf{a}_{t_a})$, $\mathbf{t}=(t_v,t_a)$,
and $\mathbf{h}^{(k)}=(\mathbf{h}_v^{(k)},
\mathbf{h}_a^{(k)})$ denote the joint inputs,
diffusion timesteps, and intermediate representations,
respectively, with $k=0,\ldots,N-1$.
The operators $\mathcal{P}_{\theta_p}$,
$\mathcal{B}_{\theta_b}$, and
$\mathcal{Q}_{\theta_q}$ represent the Pre-loop,
In-loop, and Post-loop, respectively. Each operator
incorporates joint self-attention between video
and action tokens while maintaining separate
modality-specific parameters. The In-loop parameters
$\theta_b$ are reused across all recurrent iterations,
and $N$ denotes the loop depth.

Increasing $N$ extends the effective computational
depth without increasing the number of model
parameters. This recurrent computation is performed
within each denoising step, before the sampler
advances to the next noise level, allowing the loop
depth to be controlled independently of the number
of denoising steps.

\textbf{Inference.} As in Fast-WAM~\cite{yuan2026fastwam}, future video generation is omitted during inference. We encode the observation with a predefined loop depth \(N_v\) and extract the layer-wise logical key-value (KV) features:

\[
\mathcal{C}_v =
\operatorname{ExtractKV}
\left(
\mathbf{o};N_v
\right),
\]

where \(\mathcal{C}_v\) denotes the extracted visual KV features, and \(N_v\) specifies the predefined loop depth used for feature extraction. These cached representations are subsequently reused by the action expert throughout the action denoising process, providing visual context for action generation without repeatedly encoding the observation.

\subsection{Deep Supervision}
\label{sec:deep_supervise}

To supervise predictions at different loop depths,
we apply deep supervision~\cite{lee2015deeply,szegedy2015going} to intermediate outputs
in addition to the final prediction. As in the shared
Post-loop supervision of Looped-DiT~\cite{chng2026loopeddit},
the same decoder is used across loop depths. Here,
the shared Post-loop module produces video and action predictions
from the joint hidden representations at different
loop depths:
\begin{equation}
\begin{aligned}
\hat{\mathbf{v}}^{(k)}
    &= \mathcal{Q}_{\theta_q}
       \bigl(\mathbf{h}^{(k)}\bigr)\\
    &= \bigl(\hat{\mathbf{v}}_v^{(k)},
             \hat{\mathbf{v}}_a^{(k)}\bigr).
\end{aligned}
\label{eq:intermediate_prediction}
\end{equation}
The same Post-loop parameters are reused across
all supervised depths, with
$\hat{\mathbf{v}}^{(N)}=\hat{\mathbf{v}}$.

For each modality $m\in\{v,a\}$, we apply the
flow-matching loss~\cite{lipman2023flow,liu2023flow} at depth $k$:
\begin{equation}
\mathcal{L}_m^{(k)}
=
\mathbb{E}\left[
\left\|
\hat{\mathbf{v}}_m^{(k)}
-
(\boldsymbol{\epsilon}_m-\mathbf{y}_m)
\right\|_2^2
\right],
\label{eq:intermediate_loss}
\end{equation}
where $\mathbf{y}_v=\mathbf{z}$ and
$\mathbf{y}_a=\mathbf{a}$ are the clean video
and action targets, respectively, and
$\boldsymbol{\epsilon}_m$ denotes the
corresponding Gaussian noise.

Given a total loop depth $N$, we supervise the
first $n$ intermediate predictions ($1\leq n<N$)
alongside the final prediction. The intermediate
losses are averaged, while the final loss retains
its full weight, using the final-plus-mean weighting
also used in Looped-DiT~\cite{chng2026loopeddit}:
\begin{equation}
\mathcal{L}_m
=
\mathcal{L}_m^{(N)}
+
\frac{1}{n}
\sum_{k=1}^{n}
\mathcal{L}_m^{(k)},
\label{eq:deep_supervision_objective}
\end{equation}
The overall training objective is
\begin{equation}
\mathcal{L}_{\mathrm{total}}
=
\mathcal{L}_a
+
\lambda\mathcal{L}_v,
\label{eq:deep_supervision_total}
\end{equation}
where $\lambda$ balances the action and video
losses. Since intermediate predictions reuse the
same Post-loop module and are only computed during
training, deep supervision introduces no additional
model parameters or inference-time computation.

\begin{table*}[t]
\centering

\caption{
\textbf{Success rates (\%) on the LIBERO-Plus benchmark across seven distribution-shift categories.} P.T.\ denotes embodied pretraining. Fixed-loop variants have a fixed loop depth of four. Results are zero-shot evaluations of LIBERO-trained checkpoints.
}
\label{tab:liberoplus_main}
\small
\setlength{\tabcolsep}{5pt}
\resizebox{\textwidth}{!}{
\begin{tabular}{lcccccccccc}
\toprule
Models & P.T. & Params (B)
& Camera $\uparrow$ & Robot $\uparrow$ & Language $\uparrow$
& Light $\uparrow$ & Background $\uparrow$ & Noise $\uparrow$
& Layout $\uparrow$ & Avg. $\uparrow$ \\
\midrule
$\pi_{0}$~\cite{black2024pi0}
& \cmark & 3.3 & 13.8 & 6.0 & 58.8 & 85.0 & 81.4 & 79.0 & 68.9 & 56.1 \\
$\pi_{0}$-FAST~\cite{pertsch2025fast}
& \cmark & 3.3 & \textbf{65.1} & 21.6 & 61.0 & 73.2 & 73.2 & 74.4 & 68.8 & 61.6 \\
OpenVLA-OFT~\cite{kim2025openvlaoft}
& \cmark & 7.0 & 56.4 & 31.9 & 79.5 & 88.7 & \textbf{93.3} & 75.8 & 74.2 & 69.6 \\
UniVLA~\cite{bu2025univla}
& \cmark & 7.0 & 1.8 & 46.2 & 69.6 & 69.0 & 81.0 & 21.2 & 31.9 & 42.9 \\
WorldVLA~\cite{cen2025worldvla}
& \xmark & 7.0 & 0.1 & 27.9 & 41.6 & 43.7 & 17.1 & 10.9 & 38.0 & 25.0 \\
LingBot-VA~\cite{li2026lingbotva}
& \cmark & 5.3 & 40.9 & \textbf{83.0} & \textbf{86.4} & 82.3 & 53.1 & 64.4 & 76.2 & 69.5 \\
Fast-WAM~\cite{yuan2026fastwam}
& \xmark & 6.0 & 16.4 & 44.5 & 68.9 & 78.2 & 53.7 & 37.7 & 60.7 & 51.4 \\
Motus~\cite{bi2026motus}
& \cmark & 8.0 & 17.9 & 43.3 & 74.1 & 54.7 & 44.9 & 26.2 & 52.0 & 44.0 \\
\midrule
\rowcolor{oursrow}
Looped-WAM-Fixed
& \xmark & 3.0 & 37.00 & 69.04 & 48.60 & \textbf{95.16} & 77.96 & 86.27 & 77.62 & 70.24 \\
\rowcolor{oursrow}
Looped-WAM-Adaptive
& \xmark & 3.0 & 43.42 & 71.25 & 46.27 & 94.97 & 76.92 & \textbf{89.70} & \textbf{77.74} & \textbf{71.47} \\
\bottomrule
\end{tabular}
}
\vspace{-0.4cm}
\end{table*}

\subsection{Adaptive Inference}
\label{sec:adaptive_inference}

To enable flexible computation during inference, we introduce a lightweight dual-head exit gate that adaptively determines the loop depth at each denoising step, following prior work on early exiting~\cite{teerapittayanon2016branchynet,xin2020deebert,schwartz2020right}. The gate takes a feature vector $\mathbf{g}^{(k)}$ constructed by concatenating the vectorized action hidden state $\mathbf{h}_a^{(k)}$, the normalized loop depth, and the noise level $\sigma_t$ at the current denoising step. For $k>1$, the feature vector additionally incorporates the hidden-state difference.
$\Delta\mathbf{h}_a^{(k)}
=\mathbf{h}_a^{(k)}-\mathbf{h}_a^{(k-1)}$.
We use two independent linear heads for the first
and subsequent iterations, respectively.
The exit probability is
\begin{equation}
p^{(k)}
=
\operatorname{sigmoid}
(\mathbf{W}_{j}\mathbf{g}^{(k)}+b_j),
\label{eq:exit_gate}
\end{equation}
where $j=1$ for $k=1$ and $j=2$ for $1<k<N$.

We train the gate with depth-based supervision derived from
action predictions at different loop depths. With the backbone
frozen, we execute all $N$ loops and obtain action predictions
at each depth through the shared Post-loop module. For each
depth $R$, we compute the flow-matching loss
$\mathcal{L}_a^{(R)}$ and determine the target exit depth
by minimizing
\begin{equation}
\begin{aligned}
J(R)
&= \mathcal{L}_a^{(R)}
+\beta\frac{R-1}{N-1},\\
R^*
&= \min\operatorname*{arg\,min}_{1\le R\le N}J(R),
\end{aligned}
\label{eq:exit_cost}
\end{equation}
where $\beta$ controls the trade-off between prediction
accuracy and computation.

The gate is supervised to continue before the target depth $R^*$ and exit at $R^*$ when $R^*<N$, with subsequent decisions excluded from training. To account for the relative benefit of exiting versus continuing, we weight each supervised decision according to the difference between the corresponding objective values:

\begin{equation}
w^{(k)}
= \min\left(
1,
\frac{\left|J(k)-\min_{R>k}J(R)\right|}{\gamma}
\right),
\label{eq:gate_weight}
\end{equation}
where $\gamma>0$ is a margin. The gate is optimized using
weighted binary cross-entropy:
\begin{equation}
\mathcal{L}_{\mathrm{gate}}
=
\frac{
\sum_{k\in\mathcal{D}} w^{(k)}
\ell_{\mathrm{BCE}}
\left(
\mathbf{W}_j\mathbf{g}^{(k)}+b_j,
y^{(k)}
\right)
}{
\sum_{k\in\mathcal{D}} w^{(k)}+\epsilon
},
\label{eq:gate_loss}
\end{equation}
where $\mathcal{D}$ contains valid gate decisions,
$y^{(k)}$ denotes the binary target, and
$\ell_{\mathrm{BCE}}$ denotes binary
cross-entropy with logits.
The supervision targets and weights
are treated as stop-gradient quantities, and only the
gate parameters are updated.

During inference, the In-loop module exits when \(p^{(k)}>\tau\) or the maximum depth \(N\) is reached. The shared Post-loop module generates predictions at the selected depth without modifying the denoising schedule, allowing each denoising step to independently determine its loop depth.

\section{Experiments}
\label{sec:experiments}

\subsection{Experimental Setup}
\label{sec:experiment_setup}

\paragraph{Implementation details.}
\label{sec:implementation_details}

We evaluate the effectiveness of the looped architecture with Looped-WAM. We distill a looped video backbone from Wan2.2-5B~\cite{wan2025wan}, consisting of Pre-loop, In-loop, and Post-loop stages with five Transformer blocks each, while reusing its text encoder and video VAE. After distillation, we introduce an action expert with the same looped architecture. The video backbone and action expert are then jointly optimized, with joint self-attention enabling cross-modal interactions.

Looped-WAM contains approximately 3B parameters with an action horizon of 32. During training, we set the maximum number of loops to four. For video processing, frames are temporally downsampled by a factor of four, yielding nine frames per chunk. Multi-camera images are concatenated into a single image before being encoded by the video VAE. Following Fast-WAM, we enable future video prediction during training and disable it at inference.

\paragraph{Benchmarks and evaluation.}

We evaluate Looped-WAM on two simulation benchmarks,
LIBERO-Plus and RoboTwin 2.0, and further validate its
effectiveness through real-world robotic manipulation
experiments.

\noindent \textbf{LIBERO-Plus.}
We evaluate the robustness of Looped-WAM on LIBERO-Plus~\cite{fei2026liberoplus}, which covers seven categories of distribution shifts: camera viewpoint, robot initial state, language instruction, lighting, background, sensor noise, and object layout. Following the official LIBERO evaluation protocol, we report per-category and average success rates under identical benchmark splits.

\noindent \textbf{RoboTwin 2.0.}
We evaluate Looped-WAM on RoboTwin 2.0~\cite{mu2025robotwin}, a challenging benchmark with over 50 bimanual manipulation tasks.
We train on RoboTwin-Full, comprising 2,500 clean-scene and 25,000 randomized-scene demonstrations.
We report average task success rates on RoboTwin-Clean and RoboTwin-Random, using 100 evaluation episodes per task under identical evaluation protocols.

\paragraph{Real-World Evaluation.}
We conduct real-world experiments on the AgileX NERO robotic platform, comparing Looped-WAM against Fast-WAM~\cite{yuan2026fastwam}.
We evaluate three manipulation tasks with 21 trials per task under identical hardware, observation, control, and action execution settings.
We report per-task and average success rates (SR, \%) using identical success evaluation criteria.

\subsection{Main Results}
\label{sec:main_results}

\subsubsection{Results on LIBERO-Plus}

We evaluate the robustness of Looped-WAM under seven distribution shifts on LIBERO-Plus, following the original LIBERO evaluation protocol~\cite{liu2023libero}. Category-wise and average success rates are reported in Table~\ref{tab:liberoplus_main}.

We compare representative VLA models and non-looped WAM baselines with fixed-loop and adaptive-loop variants of Looped-WAM. With approximately 3B parameters, Looped-WAM-Fixed achieves an average success rate of 70.24\%, compared with 51.4\% for the 6B Fast-WAM baseline. Adaptive inference further increases average success to 71.47\%, a gain of 1.23 percentage points over fixed-loop inference. The largest gains occur under camera viewpoint and sensor noise perturbations, at 6.42 and 3.43 percentage points, respectively, although success rates decrease under language, lighting, and background perturbations. These results show that the benefits of adaptive inference vary across distribution shifts.

\subsubsection{Results on RoboTwin 2.0}

We evaluate the manipulation performance of Looped-WAM on RoboTwin 2.0, with success rates under clean and randomized settings and their overall average reported in Table~\ref{tab:robotwin_main}.

We compare multiple WAM and VLA baselines with fixed-loop and adaptive-loop variants of Looped-WAM. Looped-WAM-Fixed achieves an average success rate of 92.8\% with approximately 3B parameters, compared with 91.8\% for the 6B Fast-WAM baseline. Adaptive inference achieves a comparable average success rate of 92.6\%, with a slight improvement in clean scenes and a decrease in randomized scenes. The fixed-loop model thus achieves competitive manipulation performance with fewer parameters than Fast-WAM, while the value of adaptive inference should be assessed jointly with its computational cost.

\begin{table}[t]
\centering
\caption{
\textbf{Comparison of success rates (\%) on RoboTwin 2.0 under clean and randomized settings.} The reported ``Looped-WAM-Fixed'' uses a fixed loop depth of four to match the computational cost of Fast-WAM. P.T.\ denotes embodied pretraining. Avg.\ denotes the arithmetic mean of the clean and randomized success rates, computed before rounding.
}
\label{tab:robotwin_main}
\small
\setlength{\tabcolsep}{5pt}
\resizebox{\columnwidth}{!}{
\begin{tabular}{lccccc}
\toprule
Models & P.T. & Params (B) & Clean$\uparrow$ & Rand.$\uparrow$ & Avg.$\uparrow$ \\
\midrule
$\pi_{0}$~\cite{black2024pi0}
& \cmark & 3.3 & 65.9 & 58.4 & 62.2 \\
$\pi_{0.5}$~\cite{black2025pi05}
& \cmark & 3.3 & 82.7 & 76.8 & 79.8 \\
StarVLA-OFT~\cite{community2026starvla}
& \xmark & 4.0 & 88.2 & 88.3 & 88.3 \\
LingBot-VA~\cite{li2026lingbotva}
& \cmark & 5.3 & 92.9 & 91.6 & 92.3 \\
Motus~\cite{bi2026motus}
& \cmark & 8.0 & 88.7 & 87.0 & 87.8 \\
GigaWorld-Policy~\cite{ye2026gigaworld}
& \cmark & 5.0 & 87.0 & 85.0 & 86.0 \\
LaWAM~\cite{chen2026lawam}
& \cmark & 2.56 & 92.64 & 89.80 & 91.22 \\
4D-WAM~\cite{yang20264dwam}
& \xmark & 6.02 & 91.92 & 90.76 & 91.34 \\
X-WAM~\cite{guo2026xwam}
& \cmark & 6.67 & 89.8 & 90.7 & 90.25 \\
Fast-WAM~\cite{yuan2026fastwam}
& \xmark & 6.0 & 91.9 & 91.8 & 91.8 \\
\midrule
\rowcolor{oursrow}
Looped-WAM-Fixed
& \xmark & 3.0 & 93.0 & \textbf{92.5} & \textbf{92.8} \\
\rowcolor{oursrow}
Looped-WAM-Adaptive
& \xmark & 3.0 & 93.1 & 92.0 & 92.6 \\
\bottomrule
\end{tabular}
}
\end{table}

\subsubsection{Real-World Performance}
\begin{figure}[t]
\centering
\includegraphics[width=\columnwidth]{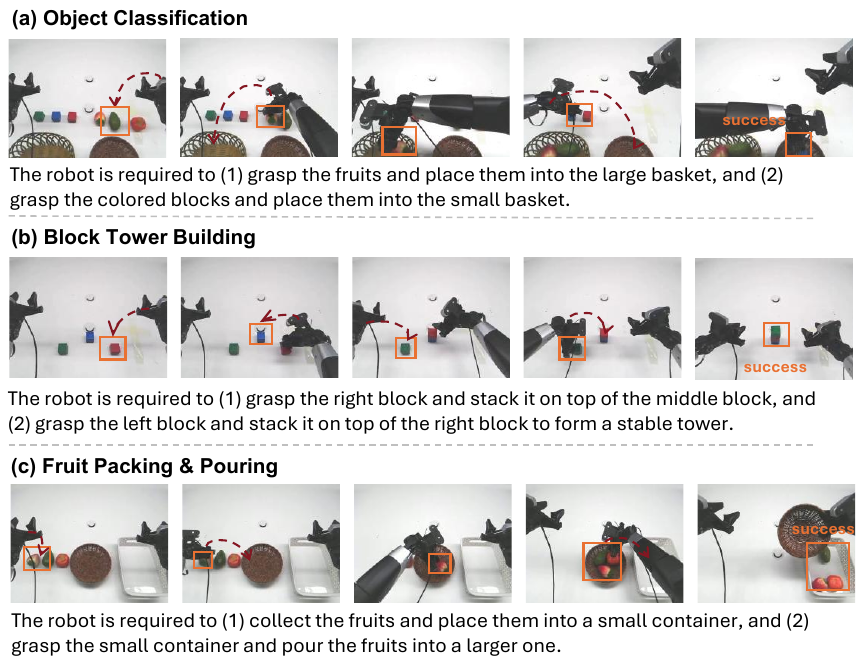}
\caption{
\textbf{Real-world rollouts of Looped-WAM on the AgileX NERO platform.} Panels (a), (b), and (c) show \textit{Object Classification}, \textit{Block Tower Building}, and \textit{Fruit Packing and Pouring}, respectively. Each row shows five chronological keyframes from left to right, with successful completion marked in the final frame.
}
\label{fig:real_world}
\end{figure}
We evaluate Looped-WAM on the AgileX NERO robotic platform and compare it with the Fast-WAM baseline~\cite{yuan2026fastwam}. Our evaluation includes three manipulation tasks:

\begin{itemize}
    \item \textbf{Object Classification}: Classify fruits and colored blocks into large and small baskets, respectively.
    \item \textbf{Block Tower Building}: Stack the right block on the middle block, then place the left block on top to form a stable three-block tower.
    \item \textbf{Fruit Packing and Pouring}: Collect fruits in a small container and pour them into a larger one.
\end{itemize}

To evaluate robustness, we randomize object and container positions, vary block colors, and alternate basket placements where applicable. For training, we collect 100 episodes of data for each task. We first train the model jointly on the combined data from all three tasks, followed by task-specific fine-tuning on each task. All methods are evaluated under identical hardware and execution conditions, with 21 trials per task. Representative rollouts are illustrated in Fig.~\ref{fig:real_world}.

Table~\ref{tab:real_world} reports the parameter counts and task-wise success rates of different methods.

\begin{table}[t]
\centering
\caption{
\textbf{Real-world Results.}
Classify, Packing, and Tower denote \textit{Object Classification}, \textit{Fruit Packing and Pouring}, and \textit{Block Tower Building}, respectively. Parentheses give the number of successes out of 21 trials. Avg.\ is the unweighted mean of the three task success rates, computed before rounding.
}
\label{tab:real_world}
\footnotesize\fontfamily{lmr}\selectfont
\setlength{\tabcolsep}{3pt}
\begin{tabular}{@{}lcc@{}}
\toprule
& Fast-WAM & Looped-WAM-Fixed \\
\midrule
Params (B) & 6.0 & 3.0 \\
\midrule
Classify & \textbf{29\%} (6/21) & 19\% (4/21) \\
Packing & 57\% (12/21) & \textbf{76\%} (16/21) \\
Tower & 62\% (13/21) & \textbf{71\%} (15/21) \\
\midrule
Avg. & 49\% & \textbf{56\%} \\
\bottomrule
\end{tabular}
\end{table}

\subsection{Experimental Analysis}
\label{sec:experimental_analysis}


\subsubsection{Effect of Recurrent Depth}

We investigate how recurrent depth affects manipulation performance on LIBERO-Plus and RoboTwin by varying the number of loop iterations for Looped-WAM, while keeping other configurations unchanged.

\begin{figure}[t]
    \centering
    \includegraphics[width=\columnwidth]{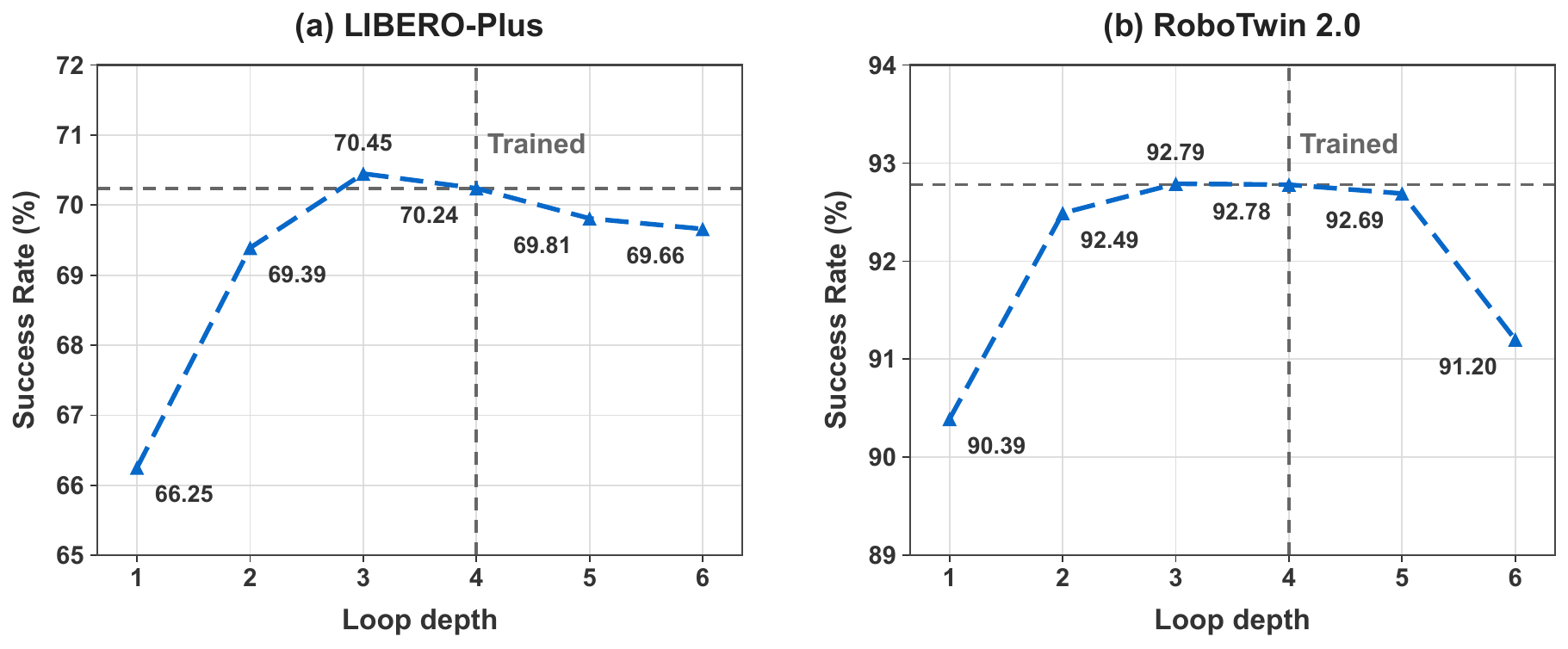}
    \caption{
    \textbf{Effect of fixed inference loop depth on average success rates on (a) LIBERO-Plus and (b) RoboTwin 2.0.} LIBERO-Plus results are averaged over seven distribution-shift categories; RoboTwin 2.0 results are averaged over clean and randomized settings. Vertical dashed lines mark the training depth of four loops, and horizontal dashed lines indicate the corresponding fixed four-loop success rates.
    }
    \label{fig:loop_depth}
\end{figure}

Under fixed-loop inference, the model executes exactly $K$ iterations, with both the video and action branches using the same fixed loop depth. In contrast, adaptive-loop inference performs a single video prefill pass, followed by adaptive looping in the action branch, with the average executed loop depth reported across evaluation episodes. Figure~\ref{fig:loop_depth} compares task success rates across different recurrent depths, highlighting the relationship between recurrent computation and manipulation performance. Table~\ref{tab:loop_depth} reports the corresponding average success rates.

Increasing the loop count from one to three improves average success from 66.25\% to 70.45\% on LIBERO-Plus and from 90.39\% to 92.79\% on RoboTwin 2.0. Further computation does not yield consistent gains: three and four loops perform nearly identically on RoboTwin 2.0, while six loops reduce success to 69.66\% and 91.20\% on the two benchmarks, respectively. Since the model is trained with a maximum of four loops, the five- and six-loop settings test extrapolation beyond the training depth. The observed decline suggests a limitation in using additional recurrent computation beyond this range, although these results alone do not establish its cause. Overall, the depth sweep distinguishes gains at shallow depths from saturation near the training depth and degradation at greater depths, motivating adaptive depth selection rather than always executing more loops.

\begin{table}[t]
\centering
\vspace{-0.3cm}
\caption{
\textbf{Success rates (\%) under different fixed loop counts $k$.}
}
\label{tab:loop_depth}
\small
\setlength{\tabcolsep}{6pt}
\scalebox{0.9}{%
\begin{tabular}{ccccc}
\toprule
\multirow{2}{*}{Loop count $k$} & \multirow{2}{*}{LIBERO-Plus} & \multicolumn{3}{c}{RoboTwin 2.0} \\
\cmidrule(lr){3-5}
& & Clean & Rand. & Avg. \\
\midrule
1 & 66.25 & 90.76 & 90.02 & 90.39 \\
2 & 69.39 & 92.84 & 92.14 & 92.49 \\
3 & \textbf{70.45} & \textbf{93.40} & 92.18 & \textbf{92.79} \\
4 & 70.24 & 93.02 & \textbf{92.54} & 92.78 \\
5 & 69.81 & 92.96 & 92.42 & 92.69 \\
6 & 69.66 & 91.72 & 90.68 & 91.20 \\
\bottomrule
\end{tabular}%
}
\vspace{-0.3cm}
\end{table}

\begin{figure}[t]
\centering
\hspace*{-0.03\columnwidth}
\includegraphics[width=1.05\columnwidth]{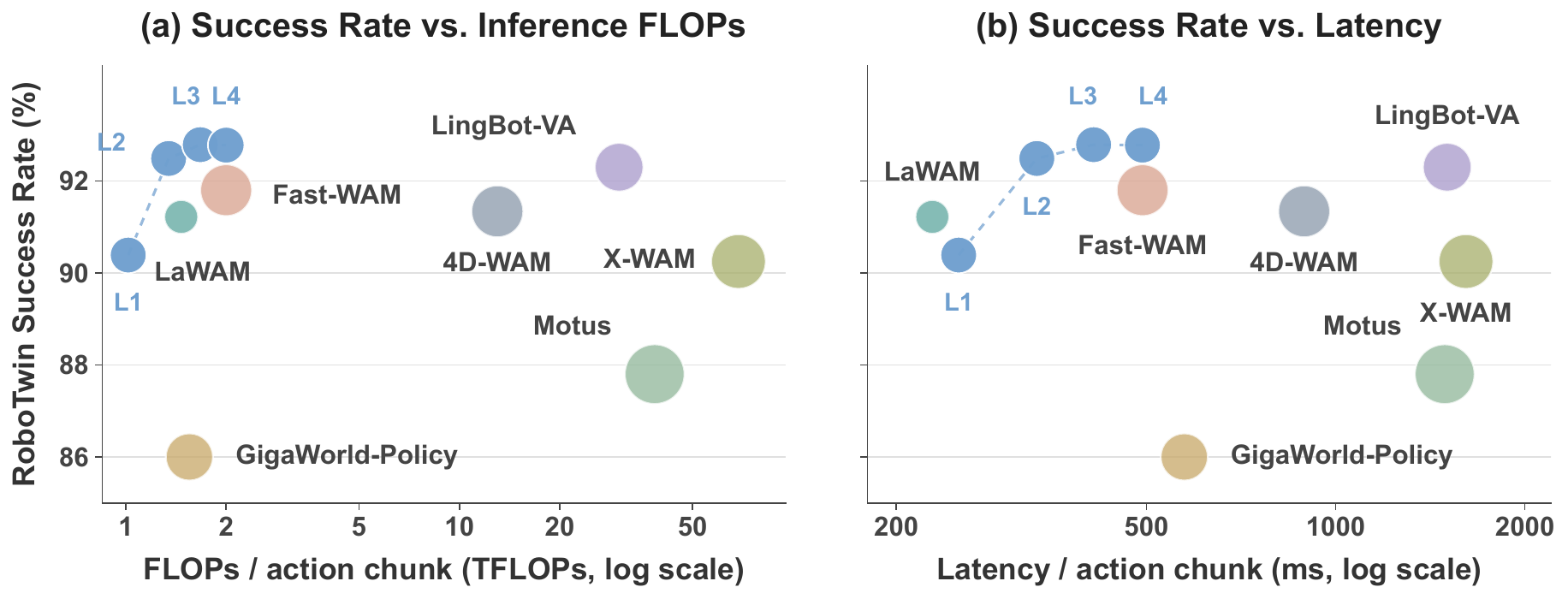}
\hspace*{0.03\columnwidth}
\vspace{-0.4cm}
\caption{
\textbf{Performance--efficiency trade-offs on RoboTwin 2.0.} Average success rates across clean and randomized settings are plotted against (a) inference FLOPs and (b) inference latency per action chunk; both horizontal axes use logarithmic scales. L1--L4 denote Looped-WAM with fixed loop counts of one to four. Circle sizes indicate model parameter counts, and colors distinguish methods. Measurements use an RTX 3090 with BF16 precision, a 16-step action chunk, and 10 denoising steps.
}
\label{fig:efficiency_tradeoff}
\vspace{-0.5cm}
\end{figure}

\subsubsection{Performance and Inference Efficiency}

We investigate the trade-off between manipulation performance and inference efficiency on RoboTwin 2.0 by examining task success rates in relation to inference FLOPs and model inference latency.

As shown in Fig.~\ref{fig:efficiency_tradeoff}, we present two comparisons: RoboTwin 2.0 task success rate versus inference FLOPs and RoboTwin 2.0 task success rate versus inference latency. The horizontal axes use logarithmic scales to highlight computational differences across model variants. Marker sizes indicate the number of model parameters, while colors and marker styles distinguish model scales and inference configurations.

All efficiency measurements are conducted on a single NVIDIA RTX 3090 GPU with 24 GB of memory, using BF16 precision, and four CPU threads. Inference FLOPs are measured for the model backbone and exclude independent encoding modules, including the video VAE and text encoder. Inference latency is measured for generating a single 16-step action chunk, including 10 denoising steps and the corresponding scheduler updates.

\begin{table*}[t]
\centering
\vspace{-0.4cm}
\caption{
\textbf{Average completion steps for five challenging RoboTwin 2.0 tasks
under different fixed loop counts $k$ and under adaptive loop
gating (Adaptive) in clean and randomized settings.}
}
\label{tab:robotwin_steps}
\resizebox{\textwidth}{!}{%
\begin{tabular}{l*{14}{c}}
\toprule
\multirow{3}{*}{Task}
& \multicolumn{14}{c}{Average Completion Steps ($\downarrow$)} \\
\cmidrule(lr){2-15}
& \multicolumn{7}{c}{Clean}
& \multicolumn{7}{c}{Random} \\
\cmidrule(lr){2-8}
\cmidrule(lr){9-15}
& $k=1$ & $k=2$ & $k=3$ & $k=4$ & $k=5$ & $k=6$ & Adaptive
& $k=1$ & $k=2$ & $k=3$ & $k=4$ & $k=5$ & $k=6$ & Adaptive \\
\midrule
Open Microwave
& 598.15 & 568.36 & 558.88 & 545.53 & 696.27 & 782.47 & \textbf{540.67}
& 557.27 & 564.76 & 558.76 & 591.24 & 653.53 & 760.46 & \textbf{539.66} \\

Put Bottles Dustbin
& 698.32 & 677.81 & 642.77 & 666.00 & 651.13 & 695.03 & \textbf{641.40}
& 693.82 & 683.59 & 647.14 & 671.47 & 633.08 & 677.04 & \textbf{627.37} \\

Blocks Ranking Size
& 607.96 & 526.97 & \textbf{470.19} & 498.96 & 524.29 & 497.52 & 514.49
& 673.62 & 583.52 & 568.61 & 550.08 & 535.14 & 541.44 & \textbf{526.50} \\

Stack Bowls Three
& 625.10 & 609.91 & 586.92 & 584.71 & 543.19 & 551.92 & \textbf{534.26}
& 665.80 & 578.60 & 563.78 & 564.36 & \textbf{536.58} & 578.13 & 571.43 \\

Hanging Mug
& 613.70 & 583.55 & 577.64 & 524.95 & 517.90 & 536.11 & \textbf{506.87}
& 609.00 & 586.16 & 572.87 & 554.18 & 597.86 & 656.02 & \textbf{548.25} \\
\bottomrule
\end{tabular}%
}
\vspace{-2mm}
\end{table*}

\subsubsection{Effect of Loop-Index Conditioning}
\label{sec:loop_index}

We further examine the impact of explicit loop-index conditioning on Looped-WAM using the LIBERO-Plus benchmark. To isolate its effect, we compare two variants that differ only in whether the current loop index is explicitly injected into the model through an adaptive layer normalization (AdaLN) layer, while keeping all other model configurations identical. Both variants are trained with a maximum of four loop iterations.

Table~\ref{tab:loop_index} presents the success rates of the two variants across seven distribution-shift categories. The results show that explicitly providing loop-index information through AdaLN reduces the average success rate, indicating that such conditioning does not improve recurrent inference in our setting. Instead, Looped-WAM achieves higher manipulation success rates without explicit loop-index conditioning, suggesting that recurrent computation can be more effective when performed without explicitly encoding the iteration index.

\begin{table}[t]
\centering
\vspace{-0.3cm}
\caption{
\textbf{Ablation study of explicit loop-index conditioning on LIBERO-Plus.}
Success rates (\%) are reported across seven distribution-shift categories.
}
\label{tab:loop_index}
\small
\setlength{\tabcolsep}{4pt}
\begin{tabular}{lcc}
\toprule
Category
& w/o Loop Index $\uparrow$
& w/ Loop Index $\uparrow$ \\
\midrule
Camera     & \textbf{37.00} & 32.58 \\
Robot      & \textbf{69.04} & 68.77 \\
Language   & \textbf{48.60} & 45.15 \\
Light      & \textbf{95.16} & 92.82 \\
Background & \textbf{77.96} & 71.28 \\
Noise      & \textbf{86.27} & 82.89 \\
Layout     & 77.62 & \textbf{77.97} \\
\midrule
Avg.       & \textbf{70.24} & 67.35 \\
\bottomrule
\end{tabular}
\vspace{-0.5cm}
\end{table}

\subsubsection{Task Completion Efficiency}

Beyond the improvements in manipulation success rates, we further observe that increasing recurrent depth can reduce the number of execution steps required to complete challenging tasks on RoboTwin 2.0. To quantify this effect, we evaluate each task over 100 episodes and measure the average number of steps needed for task completion, where fewer steps indicate higher execution efficiency.

Table~\ref{tab:robotwin_steps} presents the average number of steps required to complete five selected challenging tasks under different numbers of loop iterations. Increasing depth from one to a moderate number of loops often reduces completion steps, but the best fixed depth varies across tasks and scene settings. For example, Open Microwave in clean scenes improves from 598.15 steps at $k=1$ to 545.53 at $k=4$, before increasing to 782.47 at $k=6$. Looped-WAM-Adaptive achieves the lowest reported completion steps in eight of the ten task--setting combinations, although fixed-depth inference remains better for Blocks Ranking Size in clean scenes and Stack Bowls Three in randomized scenes. These results support flexible depth selection over a uniformly large loop count. Completion steps characterize interaction with the environment and should be considered alongside success rates and latency; fewer steps do not by themselves establish a reduction in wall-clock task completion time.

\subsection{Inference Optimization}
\label{sec:code_level_optimization}

We optimize Looped-WAM inference without modifying its architecture or trained weights. Specifically, we reuse text-conditioning embeddings and attention K/V caches across recurrent iterations and denoising steps, while eliminating redundant computations and memory operations. We further introduce Triton kernels for selected operations and employ CUDA Graphs to accelerate the inference pipeline.

\begin{figure}[t]
    \centering
    \includegraphics[width=0.9\columnwidth]{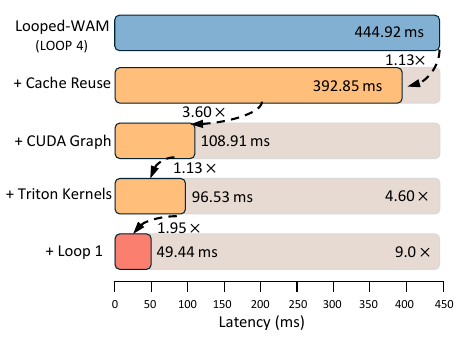}
    \vspace{-0.1cm}
    \caption{
    \textbf{Model-trunk latency with cumulative inference optimizations.} With 4 loops, cache reuse, CUDA Graph replay, and Triton kernels cut latency from 444.92 to 96.53 ms (4.60$\times$). Reducing the loop count to one brings latency to 49.44 ms (9.00$\times$ vs. baseline). Arrows show speedups between consecutive rows. All runs use identical weights and inputs on an RTX 3090 (BF16, batch size 1, 16-step action chunk, 10 denoising steps). Preprocessing, data transfers, and initial graph capture are excluded.
    }
    \label{fig:code_level_optimization}
    \vspace{-0.4cm}
\end{figure}

We evaluate inference latency on an NVIDIA RTX 3090 using BF16, a batch size of one, a 16-step action chunk, and 10 denoising steps. Measurements focus on the model trunk, excluding preprocessing, encoding, data transfers, and initial graph capture. As shown in Fig.~\ref{fig:code_level_optimization}, these optimizations achieve a 4.60$\times$ speedup without training.

\section{Conclusion}
\label{sec:conclusion}

In this paper, we presented Looped-WAM, a parameter-efficient World Action Model that introduces recurrent parameter sharing into both the video backbone and the action expert. By decoupling computational depth from model size, Looped-WAM provides a flexible framework for exploring iterative computation in robotic manipulation. Experiments on LIBERO-Plus and RoboTwin 2.0 demonstrate the potential of recurrent computation to balance manipulation performance and model efficiency.

Nevertheless, our findings also reveal several limitations. Performance gains from additional recurrent iterations can plateau or diminish, and explicit loop-index conditioning showed no clear benefit in our experiments. These findings suggest that effective recurrence depends on architectural and training choices. Future work could explore better supervision, adaptive compute allocation, and more effective use of intermediate representations to clarify the potential and limits of recurrence in efficient World Action Models.


{
    \small
    \bibliographystyle{ieeenat_fullname}
    \bibliography{main}
}


\clearpage
\appendix
\maketitlesupplementary
\raggedbottom

\section*{Supplementary Contents}
\begingroup
\small
\newcommand{\suppsectionentry}[1]{%
  \noindent\hyperref[#1]{\makebox[1.9em][l]{\ref*{#1}}\nameref*{#1}}%
  \nobreak\hspace{0.4em}\dotfill\hspace{0.4em}\pageref{#1}\par
}
\suppsectionentry{sec:supp_implementation}
\suppsectionentry{sec:supp_libero_plus}
\suppsectionentry{sec:supp_robotwin}
\suppsectionentry{sec:supp_inference_cost}
\endgroup

\section{Detailed Training Settings}
\label{sec:supp_implementation}

Looped-WAM is trained in two stages. We first initialize the looped video backbone by slicing layers from the Wan2.2-5B checkpoint~\cite{wan2025wan}, and then distill the initialized backbone, while reusing the original text encoder and video VAE. In the second stage, we jointly train the video backbone with an action expert through joint self-attention. Tables~\ref{tab:supp_distillation} and~\ref{tab:supp_training} list the hyperparameters for distillation and joint training, respectively. Both branches use five Pre-loop, five shared In-loop, and five Post-loop blocks. The model has approximately 3B parameters and a hidden dimension of 1024. Multi-view images are concatenated before VAE encoding. Temporal downsampling by a factor of four yields nine frames per video chunk.

\begin{table}[H]
\centering
\caption{
{\fontfamily{lmr}\selectfont\textbf{Hyperparameters for video backbone distillation.}}
}
\label{tab:supp_distillation}
\footnotesize
\setlength{\tabcolsep}{4pt}
\begin{tabular}{@{}p{0.60\columnwidth}p{0.35\columnwidth}@{}}
\toprule
Parameter & Value \\
\midrule
Pre / In / Post blocks & 5 / 5 / 5 \\
Training steps & 80,000 \\
Effective global batch size & 128 \\
Optimizer & AdamW \\
Learning rate & 0.001 \\
Warmup steps & 2,000 \\
LR schedule & Cosine \\
Weight decay & 0.001 \\
Gradient clipping & 1.0 \\
Precision & BF16 \\
GPU model & NVIDIA A100 \\
GPU count & 8 \\
\bottomrule
\end{tabular}
\end{table}

\newpage
\begin{table}[H]
\centering
\caption{
{\fontfamily{lmr}\selectfont\textbf{Hyperparameters for joint training.}}
}
\label{tab:supp_training}
\footnotesize
\setlength{\tabcolsep}{4pt}
\begin{tabular}{@{}p{0.60\columnwidth}p{0.35\columnwidth}@{}}
\toprule
Parameter & Value \\
\midrule
Max loop depth $N$ & 4 \\
Action horizon & 32 \\
Training epochs & 5 \\
Effective global batch size & 128 \\
Optimizer & AdamW \\
Learning rate & 0.001 \\
Warmup steps & 2,000 \\
LR schedule & Cosine \\
Weight decay & 0.001 \\
Gradient clipping & 1.0 \\
Supervised depths & $k=1,2,3,4$ \\
Video loss weight $\lambda$ & 1 \\
Precision & BF16 \\
GPU model & NVIDIA A100 \\
GPU count & 8 \\
\bottomrule
\end{tabular}
\end{table}

\section{Detailed Results on LIBERO-Plus}
\label{sec:supp_libero_plus}

\subsection{Fixed and Adaptive Inference}

Table~\ref{tab:supp_libero_tasks} reports success rates for 40 LIBERO-Plus~\cite{fei2026liberoplus} tasks across four suites. We average the seven distribution shifts equally for each task, then average the 40 tasks equally. Evaluation uses the official episode limits of 220, 280, 300, and 520 steps for Spatial, Object, Goal, and Long, respectively. 

Across fixed depths, $k=3$ has the highest average success rate (70.45\%). Adaptive inference reaches 71.47\% with a mean executed depth of 2.59, a maximum of four loops, and an exit threshold of $\tau=0.5$.

\onecolumn
\begin{table}[H]
\centering
\caption{
{\fontfamily{lmr}\selectfont\textbf{Per-task success rates (\%) on LIBERO-Plus under fixed and adaptive inference.}}
Each task gives equal weight to the seven distribution shifts. $k$ denotes the fixed loop count. Adaptive denotes Looped-WAM-Adaptive with at most four loops and exit threshold $\tau=0.5$.
}
\label{tab:supp_libero_tasks}
\footnotesize\fontfamily{lmr}\selectfont
\setlength{\tabcolsep}{4pt}
\begin{tabular}{@{}ll*{8}{r}r@{}}
\toprule
& & \multicolumn{8}{c}{Fixed} & \multirow{2}{*}{Adaptive} \\
\cmidrule(lr){3-10}
ID & Source task & $k=1$ & $k=2$ & $k=3$ & $k=4$ & $k=5$ & $k=6$ & $k=7$ & $k=8$ & \\
\midrule
\multicolumn{11}{l}{\textit{LIBERO-Spatial}} \\
S0 & Black bowl between plate and ramekin & 87.36 & 85.23 & \textbf{90.44} & 85.68 & 89.30 & 89.18 & 87.57 & 87.93 & 85.39 \\
S1 & Black bowl next to ramekin & 60.47 & 63.56 & 66.80 & \textbf{70.65} & 67.20 & 67.02 & 65.85 & 65.48 & 68.30 \\
S2 & Black bowl at table center & 76.58 & 84.00 & \textbf{86.18} & 84.08 & 84.51 & \textbf{86.18} & 84.95 & 84.45 & 81.08 \\
S3 & Black bowl on cookie box & 74.82 & \textbf{76.55} & 75.70 & 72.84 & 73.71 & 73.23 & 75.81 & 75.58 & 67.10 \\
S4 & Black bowl in top cabinet drawer & 64.27 & 65.42 & 68.23 & 69.20 & 67.25 & 62.92 & 59.65 & 54.01 & \textbf{76.59} \\
S5 & Black bowl on ramekin & 69.84 & \textbf{75.29} & 74.89 & 71.70 & 73.23 & 73.70 & 72.53 & 72.41 & 67.71 \\
S6 & Black bowl next to cookie box & 87.90 & 87.97 & 91.22 & 90.39 & \textbf{91.54} & 90.70 & 88.83 & 89.88 & 91.09 \\
S7 & Black bowl on stove & 61.83 & 68.26 & 66.67 & 66.52 & 66.46 & 63.77 & 64.55 & 64.34 & \textbf{70.77} \\
S8 & Black bowl next to plate & 74.04 & 71.57 & 70.66 & 71.45 & 69.96 & 73.71 & 74.93 & 77.01 & \textbf{79.97} \\
S9 & Black bowl on wooden cabinet & 70.85 & 74.98 & 75.66 & 74.59 & 74.28 & 72.05 & 72.70 & 72.63 & \textbf{77.09} \\
\midrule
\multicolumn{11}{l}{\textit{LIBERO-Object}} \\
O0 & Alphabet soup into basket & 82.62 & 83.75 & \textbf{88.09} & 85.41 & 83.85 & 85.99 & 84.52 & 84.47 & 87.57 \\
O1 & Cream cheese into basket & 79.00 & \textbf{85.02} & 82.56 & 83.46 & 81.47 & 84.16 & 80.73 & 81.51 & 80.64 \\
O2 & Salad dressing into basket & 80.78 & \textbf{81.11} & 76.26 & 76.66 & 77.53 & 74.75 & 76.16 & 75.35 & 78.95 \\
O3 & BBQ sauce into basket & 65.48 & 67.52 & 69.04 & 71.37 & 71.50 & 71.25 & \textbf{73.01} & 69.82 & 67.13 \\
O4 & Ketchup into basket & 45.37 & 58.67 & \textbf{63.60} & 56.92 & 57.88 & 58.76 & 57.27 & 57.34 & 57.99 \\
O5 & Tomato sauce into basket & 69.08 & 72.17 & 71.94 & 69.90 & 66.78 & 68.66 & 67.17 & 66.80 & \textbf{72.52} \\
O6 & Butter into basket & 69.71 & 73.76 & 74.80 & 75.92 & 73.16 & 74.30 & 75.26 & 76.73 & \textbf{79.36} \\
O7 & Milk into basket & 60.94 & 65.91 & 64.02 & 64.66 & 65.44 & 64.75 & 65.84 & 60.91 & \textbf{72.98} \\
O8 & Chocolate pudding into basket & 77.65 & 82.07 & 82.86 & 83.23 & 83.72 & 81.65 & 81.98 & 81.26 & \textbf{86.87} \\
O9 & Orange juice into basket & 69.20 & 70.64 & 70.78 & 71.46 & 71.65 & \textbf{71.72} & 71.18 & 69.71 & 70.99 \\
\midrule
\multicolumn{11}{l}{\textit{LIBERO-Goal}} \\
G0 & Open middle cabinet drawer & 64.93 & 74.44 & 74.89 & 75.22 & 75.22 & 74.89 & 74.23 & 73.24 & \textbf{80.68} \\
G1 & Place bowl on stove & 73.69 & 72.42 & 72.79 & 74.39 & 74.44 & 73.99 & \textbf{75.33} & 71.17 & 74.44 \\
G2 & Place wine bottle atop cabinet & 69.75 & 71.20 & \textbf{71.91} & 70.83 & 70.83 & 69.69 & 70.07 & 67.87 & 71.19 \\
G3 & Open top drawer and insert bowl & 56.23 & 59.28 & 60.70 & \textbf{61.88} & 59.66 & 60.55 & 57.98 & 58.90 & 59.69 \\
G4 & Place bowl atop cabinet & 66.74 & 66.85 & 69.66 & 68.36 & 69.91 & 68.42 & 69.64 & 67.62 & \textbf{70.95} \\
G5 & Push plate to front of stove & 54.43 & 54.96 & 54.01 & 54.74 & 55.40 & 55.44 & 54.35 & 53.71 & \textbf{55.70} \\
G6 & Put cream cheese in bowl & 61.68 & 63.05 & 61.57 & 62.73 & \textbf{64.22} & 62.88 & 61.52 & 63.93 & 62.50 \\
G7 & Turn on stove & 79.37 & 82.87 & 80.73 & 79.03 & 78.64 & 80.62 & 82.01 & 82.08 & \textbf{83.33} \\
G8 & Place bowl on plate & 73.91 & 80.16 & 80.30 & 79.87 & 79.40 & \textbf{81.08} & 81.06 & 80.63 & 78.71 \\
G9 & Place wine bottle on rack & 56.13 & 62.12 & \textbf{63.92} & 62.45 & 58.72 & 60.88 & 60.07 & 55.49 & 62.20 \\
\midrule
\multicolumn{11}{l}{\textit{LIBERO-Long}} \\
L0 & Soup and tomato sauce into basket & 74.18 & 75.57 & 78.45 & 76.92 & 76.67 & 72.96 & 71.20 & 71.14 & \textbf{81.52} \\
L1 & Cream cheese and butter into basket & 70.70 & 71.82 & 72.51 & 71.61 & 72.08 & 72.24 & 73.75 & 72.51 & \textbf{75.85} \\
L2 & Turn on stove and place moka pot & 75.31 & 78.27 & \textbf{79.05} & 77.64 & 78.61 & 76.73 & 76.56 & 76.63 & 76.92 \\
L3 & Put bowl in bottom drawer and close & 61.74 & 61.50 & 64.64 & 67.16 & 65.31 & 64.02 & 63.64 & 65.08 & \textbf{69.03} \\
L4 & Put two mugs on separate plates & 39.50 & 40.33 & 41.17 & 43.86 & 42.80 & 42.81 & 42.09 & 43.60 & \textbf{45.17} \\
L5 & Place book in back caddy compartment & 75.16 & 76.98 & 75.57 & 79.15 & 77.91 & 78.41 & \textbf{82.51} & 81.91 & 78.85 \\
L6 & Place mug and pudding beside plate & 33.03 & 47.04 & 45.96 & 46.61 & 45.58 & 46.53 & 47.36 & 45.88 & \textbf{50.38} \\
L7 & Soup and cream cheese into basket & 62.62 & 64.26 & 68.28 & 66.94 & 66.06 & 65.72 & 62.44 & 65.77 & \textbf{71.10} \\
L8 & Put both moka pots on stove & 34.76 & 34.09 & 39.85 & 42.30 & 39.66 & 39.27 & 42.38 & 41.72 & \textbf{48.67} \\
L9 & Put mug in microwave and close & 38.44 & 44.99 & 51.46 & \textbf{51.69} & 50.77 & 50.96 & 45.92 & 46.34 & 41.71 \\
\midrule
\multicolumn{2}{l}{Average} & 66.25 & 69.39 & 70.45 & 70.24 & 69.81 & 69.66 & 69.36 & 68.82 & \textbf{71.47} \\
\bottomrule
\end{tabular}
\end{table}

\clearpage
\subsection{Average Episode Steps}
\label{sec:supp_libero_steps}

Table~\ref{tab:supp_libero_steps_by_category} reports mean episode steps by distribution shift for fixed depths $k=1,\ldots,8$. The means include all evaluated episodes. Episodes that time out contribute the suite-specific step limit.

\begin{table}[H]
\centering
\caption{
{\fontfamily{lmr}\selectfont\textbf{Average episode steps on LIBERO-Plus by distribution shift and fixed loop count $k$.}}
Each category pools its episodes. Overall pools all 10,030 episodes at each loop count.
Bold marks the fewest steps in each category.
}
\label{tab:supp_libero_steps_by_category}
\footnotesize\fontfamily{lmr}\selectfont
\setlength{\tabcolsep}{4pt}
\begin{tabular}{@{}l*{8}{r}@{}}
\toprule
Category & $k=1$ & $k=2$ & $k=3$ & $k=4$ & $k=5$ & $k=6$ & $k=7$ & $k=8$ \\
\midrule
Camera Viewpoints & 295.98 & 289.49 & \textbf{288.29} & 288.81 & 289.41 & 290.99 & 290.76 & 293.00 \\
Robot Initial States & 224.99 & 219.57 & 217.91 & \textbf{217.64} & 217.65 & 218.25 & 219.46 & 220.70 \\
Language Instructions & 263.95 & 258.89 & 256.88 & \textbf{255.52} & 256.66 & 256.96 & 257.60 & 259.16 \\
Light Conditions & 181.45 & 168.71 & \textbf{166.02} & 167.58 & 167.43 & 167.07 & 168.65 & 167.14 \\
Background Textures & 221.01 & 217.77 & 218.51 & 218.17 & 219.01 & \textbf{217.46} & 220.62 & 222.46 \\
Sensor Noise & 202.45 & 195.50 & 193.12 & \textbf{191.64} & 193.14 & 194.24 & 194.61 & 195.19 \\
Objects Layout & 193.31 & 188.81 & 188.29 & \textbf{187.65} & 188.21 & 189.03 & 190.51 & 190.82 \\
\midrule
\textbf{Overall} & 228.48 & 222.24 & 220.80 & \textbf{220.44} & 221.11 & 221.59 & 222.64 & 223.60 \\
\bottomrule
\end{tabular}
\end{table}

\subsection{Layer-Sampled Initialization from OpenWAM}
\label{sec:supp_libero_openwam}

To further assess the generality of the looped architecture, we initialize the video backbone from an OpenWAM~\cite{wang2026openwam} checkpoint pretrained on LIBERO. We select 15 layers from the OpenWAM video backbone via stratified sampling and randomly initialize the action expert with the same Pre-loop, In-loop, and Post-loop structure. We then train both branches jointly. The resulting model reaches an average success rate of 74.00\% on LIBERO-Plus (Table~\ref{tab:supp_libero_openwam_init}). Together with the results for the distilled backbone in Table~\ref{tab:supp_libero_tasks}, these results show that the looped architecture performs well with both distillation and layer-sampled initialization.

\begin{table}[H]
\centering
\caption{
{\fontfamily{lmr}\selectfont\textbf{Success rates (\%) on LIBERO-Plus with layer-sampled initialization from OpenWAM.}} We evaluate the epoch-5 checkpoint at a fixed loop depth of \(k=4\). To compute the overall success rate, we pool evaluation episodes across all 40 tasks and report the proportion of successful episodes.
}
\label{tab:supp_libero_openwam_init}
\footnotesize\fontfamily{lmr}\selectfont
\setlength{\tabcolsep}{2pt}
\begin{minipage}[t]{0.485\textwidth}
\begin{tabular*}{\linewidth}{@{\extracolsep{\fill}}llr@{}}
\toprule
ID & Source task & SR (\%) \\
\midrule
\multicolumn{2}{@{}l}{\textbf{LIBERO-Spatial} (avg.)} & \textbf{86.05} \\
S0 & Black bowl between plate and ramekin & 86.33 \\
S1 & Black bowl next to ramekin & 91.84 \\
S2 & Black bowl at table center & 87.43 \\
S3 & Black bowl on cookie box & 91.02 \\
S4 & Black bowl in top cabinet drawer & 76.86 \\
S5 & Black bowl on ramekin & 83.70 \\
S6 & Black bowl next to cookie box & 79.86 \\
S7 & Black bowl on stove & \textbf{94.21} \\
S8 & Black bowl next to plate & 79.78 \\
S9 & Black bowl on wooden cabinet & 89.53 \\
\midrule
\multicolumn{2}{@{}l}{\textbf{LIBERO-Object} (avg.)} & \textbf{76.18} \\
O0 & Alphabet soup into basket & 77.72 \\
O1 & Cream cheese into basket & 70.41 \\
O2 & Salad dressing into basket & \textbf{86.01} \\
O3 & BBQ sauce into basket & 71.01 \\
O4 & Ketchup into basket & 69.87 \\
O5 & Tomato sauce into basket & 79.32 \\
O6 & Butter into basket & 69.08 \\
O7 & Milk into basket & 78.25 \\
O8 & Chocolate pudding into basket & 76.34 \\
O9 & Orange juice into basket & 83.82 \\
\bottomrule
\end{tabular*}
\end{minipage}\hfill
\begin{minipage}[t]{0.485\textwidth}
\begin{tabular*}{\linewidth}{@{\extracolsep{\fill}}llr@{}}
\toprule
ID & Source task & SR (\%) \\
\midrule
\multicolumn{2}{@{}l}{\textbf{LIBERO-Goal} (avg.)} & \textbf{69.74} \\
G0 & Open middle cabinet drawer & 61.14 \\
G1 & Place bowl on stove & 79.22 \\
G2 & Place wine bottle atop cabinet & 79.24 \\
G3 & Open top drawer and insert bowl & 57.03 \\
G4 & Place bowl atop cabinet & 72.74 \\
G5 & Push plate to front of stove & 53.62 \\
G6 & Put cream cheese in bowl & 66.90 \\
G7 & Turn on stove & 81.47 \\
G8 & Place bowl on plate & \textbf{84.16} \\
G9 & Place wine bottle on rack & 61.84 \\
\midrule
\multicolumn{2}{@{}l}{\textbf{LIBERO-Long} (avg.)} & \textbf{74.31} \\
L0 & Soup and tomato sauce into basket & 88.73 \\
L1 & Cream cheese and butter into basket & \textbf{89.75} \\
L2 & Turn on stove and place moka pot & 74.23 \\
L3 & Put bowl in bottom drawer and close & 72.76 \\
L4 & Put two mugs on separate plates & 71.85 \\
L5 & Place book in back caddy compartment & 77.52 \\
L6 & Place mug and pudding beside plate & 57.01 \\
L7 & Soup and cream cheese into basket & 79.18 \\
L8 & Put both moka pots on stove & 68.87 \\
L9 & Put mug in microwave and close & 63.17 \\
\bottomrule
\end{tabular*}
\end{minipage}
\par\smallskip{\normalsize\textbf{Average success rate: 74.00\%}\par}
\par\smallskip\noindent\rule{\textwidth}{\heavyrulewidth}
\end{table}

\clearpage
\section{Detailed Results on RoboTwin 2.0}
\label{sec:supp_robotwin}

We report per-task success rates on the 50 RoboTwin~2.0 tasks under Clean and Randomized settings. We evaluate 100 episodes for each task and setting. Table~\ref{tab:supp_robotwin_l} reports Looped-WAM-Fixed for $k=1,\ldots,8$ and Looped-WAM-Adaptive. For Looped-WAM-Adaptive, the average number of loops per episode is 2.47 under the Clean setting and 2.39 under the Randomized setting. Each setting averages over 5,000 episodes.

\begin{table}[H]
\centering
\footnotesize\fontfamily{lmr}\selectfont
\setlength{\tabcolsep}{1.05pt}
\caption{
{\fontfamily{lmr}\selectfont\textbf{Per-task success rates (\%) on RoboTwin~2.0 in Clean and Randomized settings.}}
Numbers in the column headers denote fixed loop counts. Adap.\ denotes Looped-WAM-Adaptive.
}
\label{tab:supp_robotwin_l}
\begin{tabular}{@{}l@{\hspace{4pt}}*{8}{c}c@{\hspace{4pt}}*{8}{c}c@{}}
\toprule
& \multicolumn{9}{c}{Clean} & \multicolumn{9}{c}{Randomized} \\
\cmidrule(lr){2-10}\cmidrule(lr){11-19}
Task & $1$ & $2$ & $3$ & $4$ & $5$ & $6$ & $7$ & $8$ & Adap.
& $1$ & $2$ & $3$ & $4$ & $5$ & $6$ & $7$ & $8$ & Adap. \\
\midrule
Adjust Bottle & \textbf{100} & \textbf{100} & \textbf{100} & \textbf{100} & \textbf{100} & \textbf{100} & \textbf{100} & \textbf{100} & \textbf{100.0} & \textbf{100} & \textbf{100} & \textbf{100} & \textbf{100} & \textbf{100} & \textbf{100} & 99 & \textbf{100} & \textbf{100.0} \\
Beat Block Hammer & 99 & 99 & 99 & \textbf{100} & \textbf{100} & 98 & 92 & 92 & 97.0 & 91 & 95 & 91 & 93 & 93 & \textbf{96} & 81 & 67 & 91.0 \\
Blocks Ranking RGB & 97 & 98 & \textbf{99} & 97 & \textbf{99} & 98 & \textbf{99} & 97 & 98.0 & 94 & 94 & \textbf{95} & \textbf{95} & \textbf{95} & \textbf{95} & \textbf{95} & 94 & 94.0 \\
Blocks Ranking Size & 81 & 91 & \textbf{97} & 93 & 89 & 93 & 90 & 81 & 90.0 & 76 & 84 & 88 & 89 & \textbf{92} & 90 & 81 & 71 & 86.0 \\
Click Alarmclock & \textbf{100} & \textbf{100} & \textbf{100} & \textbf{100} & \textbf{100} & \textbf{100} & \textbf{100} & 99 & \textbf{100.0} & \textbf{100} & \textbf{100} & \textbf{100} & \textbf{100} & \textbf{100} & \textbf{100} & \textbf{100} & \textbf{100} & \textbf{100.0} \\
Click Bell & \textbf{100} & \textbf{100} & \textbf{100} & \textbf{100} & \textbf{100} & \textbf{100} & \textbf{100} & \textbf{100} & \textbf{100.0} & \textbf{100} & \textbf{100} & \textbf{100} & \textbf{100} & \textbf{100} & \textbf{100} & \textbf{100} & \textbf{100} & \textbf{100.0} \\
Dump Bin Bigbin & 93 & 94 & 97 & 95 & 94 & 95 & 94 & 84 & \textbf{98.0} & 96 & \textbf{98} & 93 & 96 & 97 & 92 & 94 & 82 & 96.0 \\
Grab Roller & \textbf{100} & \textbf{100} & \textbf{100} & \textbf{100} & \textbf{100} & \textbf{100} & \textbf{100} & \textbf{100} & \textbf{100.0} & \textbf{100} & \textbf{100} & \textbf{100} & \textbf{100} & \textbf{100} & \textbf{100} & \textbf{100} & \textbf{100} & \textbf{100.0} \\
Handover Block & 95 & \textbf{98} & \textbf{98} & \textbf{98} & 97 & 96 & 81 & 67 & \textbf{98.0} & 87 & 89 & 88 & 87 & \textbf{94} & 86 & 77 & 52 & 89.0 \\
Handover Mic & 99 & 97 & 97 & 98 & 98 & \textbf{100} & 99 & 98 & \textbf{100.0} & \textbf{100} & \textbf{100} & \textbf{100} & 99 & \textbf{100} & \textbf{100} & 99 & \textbf{100} & 99.0 \\
Hanging Mug & 48 & 53 & 54 & 65 & 64 & 61 & 62 & 52 & \textbf{66.0} & 49 & 53 & 55 & \textbf{59} & 51 & 41 & 51 & 30 & 56.0 \\
Lift Pot & \textbf{100} & \textbf{100} & \textbf{100} & \textbf{100} & \textbf{100} & 95 & 81 & 82 & \textbf{100.0} & 99 & \textbf{100} & \textbf{100} & \textbf{100} & \textbf{100} & 98 & 85 & 77 & \textbf{100.0} \\
Move Can Pot & 94 & 97 & 98 & \textbf{99} & \textbf{99} & 96 & 79 & 60 & \textbf{99.0} & \textbf{99} & \textbf{99} & 98 & 98 & 97 & 92 & 73 & 50 & 97.0 \\
Move Pillbottle Pad & \textbf{100} & 99 & 99 & 99 & 98 & 99 & 93 & 86 & 98.0 & \textbf{100} & \textbf{100} & \textbf{100} & \textbf{100} & \textbf{100} & 99 & 97 & 94 & \textbf{100.0} \\
Move Playingcard Away & 99 & \textbf{100} & \textbf{100} & \textbf{100} & \textbf{100} & \textbf{100} & 99 & 99 & \textbf{100.0} & 99 & \textbf{100} & \textbf{100} & \textbf{100} & \textbf{100} & \textbf{100} & \textbf{100} & \textbf{100} & \textbf{100.0} \\
Move Stapler Pad & 74 & \textbf{83} & 75 & 78 & 80 & 77 & 45 & 34 & 74.0 & 70 & \textbf{84} & 83 & 83 & 81 & 68 & 52 & 40 & \textbf{84.0} \\
Open Laptop & 93 & 94 & 95 & 95 & 95 & 96 & 94 & \textbf{97} & 93.0 & 94 & \textbf{98} & \textbf{98} & \textbf{98} & 97 & 97 & 97 & 97 & \textbf{98.0} \\
Open Microwave & 94 & \textbf{95} & 90 & 88 & 78 & 70 & 58 & 50 & 91.0 & \textbf{97} & 96 & 96 & 96 & 88 & 82 & 63 & 50 & 96.0 \\
Pick Diverse Bottles & 80 & 85 & 84 & \textbf{86} & 83 & 81 & 77 & 71 & 85.0 & 79 & 83 & 85 & \textbf{89} & \textbf{89} & 87 & 81 & 77 & 88.0 \\
Pick Dual Bottles & 93 & 96 & 96 & \textbf{99} & 97 & 95 & 75 & 66 & 97.0 & 91 & 96 & \textbf{98} & 95 & 93 & 95 & 79 & 73 & 95.0 \\
Place A2B Left & 90 & 92 & \textbf{95} & 94 & \textbf{95} & 94 & 93 & 89 & 94.0 & 89 & 91 & 92 & 91 & \textbf{93} & \textbf{93} & 90 & 88 & \textbf{93.0} \\
Place A2B Right & 90 & 92 & 89 & 91 & 91 & 90 & 92 & \textbf{94} & \textbf{94.0} & 91 & 93 & 93 & 94 & 95 & 95 & \textbf{96} & 95 & 95.0 \\
Place Bread Basket & 94 & 94 & \textbf{96} & 91 & 93 & 90 & 85 & 87 & \textbf{96.0} & \textbf{93} & \textbf{93} & 92 & 92 & \textbf{93} & 92 & 91 & 88 & \textbf{93.0} \\
Place Bread Skillet & 93 & 90 & 94 & 92 & 85 & \textbf{95} & 91 & 86 & 88.0 & 88 & 92 & 87 & 89 & 89 & \textbf{93} & 89 & 87 & 89.0 \\
Place Burger Fries & 95 & 97 & 97 & 97 & 96 & \textbf{99} & 96 & 98 & 97.0 & 95 & 94 & \textbf{96} & 95 & 95 & 95 & 93 & 94 & 93.0 \\
Place Can Basket & 70 & \textbf{75} & 74 & 69 & 74 & 63 & 50 & 47 & 67.0 & 60 & 64 & 61 & \textbf{67} & 64 & 58 & 48 & 39 & 61.0 \\
Place Cans Plasticbox & 98 & \textbf{100} & \textbf{100} & \textbf{100} & \textbf{100} & 99 & 85 & 57 & \textbf{100.0} & \textbf{99} & \textbf{99} & 96 & 98 & 97 & 97 & 94 & 69 & \textbf{99.0} \\
Place Container Plate & 98 & 99 & 99 & \textbf{100} & 95 & 98 & 98 & 99 & \textbf{100.0} & 97 & \textbf{100} & 98 & \textbf{100} & 98 & 99 & 96 & 96 & 99.0 \\
Place Dual Shoes & 92 & 91 & 89 & 91 & \textbf{94} & 92 & 79 & 69 & 90.0 & 83 & 89 & 86 & \textbf{90} & 86 & 88 & 72 & 65 & 89.0 \\
Place Empty Cup & 98 & 98 & 98 & 98 & \textbf{99} & \textbf{99} & \textbf{99} & 96 & \textbf{99.0} & 99 & 99 & \textbf{100} & \textbf{100} & \textbf{100} & \textbf{100} & \textbf{100} & 99 & \textbf{100.0} \\
Place Fan & 91 & 89 & \textbf{95} & 94 & \textbf{95} & \textbf{95} & 91 & 91 & 94.0 & 92 & 96 & 95 & 97 & \textbf{99} & 91 & 93 & 88 & 95.0 \\
Place Mouse Pad & 81 & 91 & 91 & 91 & 92 & \textbf{94} & 85 & 59 & 89.0 & 85 & 88 & 91 & 89 & 90 & \textbf{93} & 86 & 62 & 87.0 \\
Place Object Basket & 85 & 91 & 90 & 85 & 90 & \textbf{93} & 88 & 87 & 89.0 & 83 & 78 & 81 & 82 & 82 & \textbf{86} & 82 & 85 & 81.0 \\
Place Object Scale & 90 & 93 & \textbf{94} & \textbf{94} & 93 & 91 & 87 & 86 & 92.0 & 97 & \textbf{98} & 96 & 95 & 97 & 97 & \textbf{98} & 92 & 97.0 \\
Place Object Stand & 97 & 98 & \textbf{100} & 97 & 98 & 92 & 92 & 86 & 98.0 & \textbf{97} & 96 & 96 & \textbf{97} & \textbf{97} & \textbf{97} & 94 & 84 & 93.0 \\
Place Phone Stand & 94 & 96 & 97 & 96 & 96 & \textbf{98} & 95 & 92 & \textbf{98.0} & 93 & 96 & 96 & \textbf{97} & \textbf{97} & 94 & 94 & 88 & 95.0 \\
Place Shoe & 96 & \textbf{99} & \textbf{99} & \textbf{99} & 98 & \textbf{99} & 98 & 97 & 98.0 & 97 & \textbf{99} & \textbf{99} & 98 & 98 & 98 & 94 & 94 & \textbf{99.0} \\
Press Stapler & 89 & 91 & 94 & 93 & 95 & 94 & 95 & \textbf{97} & 92.0 & 90 & 89 & 92 & 92 & 92 & 91 & 92 & \textbf{93} & 92.0 \\
Put Bottles Dustbin & 89 & 91 & \textbf{94} & 90 & 93 & 89 & 90 & 72 & 93.0 & 90 & 91 & 94 & 92 & \textbf{95} & 91 & 87 & 63 & 91.0 \\
Put Object Cabinet & 91 & \textbf{92} & \textbf{92} & 89 & 90 & 84 & 89 & 83 & 88.0 & 86 & 88 & \textbf{92} & 88 & 88 & 91 & \textbf{92} & 83 & 89.0 \\
Rotate QRcode & 91 & 93 & 93 & \textbf{94} & 92 & 92 & 88 & 72 & 91.0 & 83 & 84 & 86 & \textbf{88} & 85 & 85 & 80 & 65 & 85.0 \\
Scan Object & 87 & 89 & 91 & \textbf{93} & 90 & 90 & 84 & 66 & 91.0 & 86 & \textbf{89} & 88 & 87 & 86 & 84 & 73 & 62 & 86.0 \\
Shake Bottle & \textbf{100} & \textbf{100} & \textbf{100} & \textbf{100} & \textbf{100} & \textbf{100} & 99 & \textbf{100} & \textbf{100.0} & \textbf{100} & \textbf{100} & \textbf{100} & \textbf{100} & \textbf{100} & \textbf{100} & \textbf{100} & \textbf{100} & \textbf{100.0} \\
\shortstack[l]{Shake Bottle\\Horizontally} & \textbf{100} & \textbf{100} & \textbf{100} & \textbf{100} & \textbf{100} & \textbf{100} & \textbf{100} & \textbf{100} & \textbf{100.0} & \textbf{100} & \textbf{100} & \textbf{100} & \textbf{100} & \textbf{100} & \textbf{100} & 99 & \textbf{100} & \textbf{100.0} \\
Stack Blocks Three & 97 & 98 & 98 & 98 & 98 & 98 & 97 & 87 & \textbf{99.0} & 96 & \textbf{97} & 96 & 96 & 96 & 95 & 93 & 80 & \textbf{97.0} \\
Stack Blocks Two & \textbf{100} & \textbf{100} & \textbf{100} & \textbf{100} & \textbf{100} & \textbf{100} & \textbf{100} & 98 & \textbf{100.0} & 98 & \textbf{100} & \textbf{100} & \textbf{100} & \textbf{100} & 99 & \textbf{100} & 97 & \textbf{100.0} \\
Stack Bowls Three & 76 & 78 & 80 & 81 & \textbf{86} & 85 & 68 & 77 & 84.0 & 71 & 82 & 84 & 83 & \textbf{87} & 82 & 85 & 71 & 83.0 \\
Stack Bowls Two & 91 & 96 & \textbf{98} & 94 & 95 & 94 & 96 & 96 & 96.0 & 96 & \textbf{98} & \textbf{98} & 96 & 97 & 95 & 97 & 96 & 96.0 \\
Stamp Seal & 81 & 88 & 93 & 91 & \textbf{94} & 82 & 64 & 54 & 90.0 & 87 & 92 & 93 & 94 & \textbf{95} & 87 & 72 & 64 & 94.0 \\
Turn Switch & 55 & 62 & 62 & 59 & 60 & 47 & 21 & 15 & \textbf{64.0} & 59 & \textbf{63} & \textbf{63} & \textbf{63} & \textbf{63} & 50 & 35 & 21 & 62.0 \\
\midrule
\addlinespace[1.5pt]
Average & 90.76 & 92.84 & \textbf{93.40} & 93.02 & 92.96 & 91.72 & 86.26 & 81.04 & 93.10 & 90.02 & 92.14 & 92.18 & \textbf{92.54} & 92.42 & 90.68 & 86.38 & 79.24 & 92.04 \\
\addlinespace[1.5pt]
\bottomrule
\end{tabular}
\end{table}

\clearpage
\subsection{Average Episode Steps}
\label{sec:supp_robotwin_steps}

Tables~\ref{tab:supp_robotwin_steps_clean} and~\ref{tab:supp_robotwin_steps_random} report average episode steps for Looped-WAM-Fixed ($k=1,\ldots,8$) and Looped-WAM-Adaptive. The means include successful and failed episodes. Failed episodes contribute the task-specific step limit. Table~\ref{tab:supp_robotwin_l} reports the corresponding success rates.

\begin{table}[H]
\centering
\caption{
{\fontfamily{lmr}\selectfont\textbf{Average episode steps on RoboTwin~2.0 in the Clean setting.}}
Each mean is computed over 100 episodes. $k$ denotes the fixed loop count.
Bold marks the fewest steps in each row.
}
\label{tab:supp_robotwin_steps_clean}
\footnotesize\fontfamily{lmr}\selectfont
\setlength{\tabcolsep}{4pt}
\begin{tabular}{@{}l*{9}{r}@{}}
\toprule
Task & $k=1$ & $k=2$ & $k=3$ & $k=4$ & $k=5$ & $k=6$ & $k=7$ & $k=8$ & Adaptive \\
\midrule
Adjust Bottle & 114.42 & 115.06 & 115.69 & 115.94 & 114.22 & 113.19 & 112.00 & \textbf{111.32} & 115.21 \\
Beat Block Hammer & 108.62 & 108.65 & 107.18 & 112.96 & \textbf{104.77} & 110.64 & 129.65 & 131.25 & 113.10 \\
Blocks Ranking RGB & 463.24 & 458.22 & \textbf{443.65} & 447.38 & 449.45 & 451.35 & 444.47 & 464.40 & 451.53 \\
Blocks Ranking Size & 607.96 & 526.97 & \textbf{470.19} & 498.96 & 524.29 & 497.52 & 533.92 & 595.15 & 514.49 \\
Click Alarmclock & 62.08 & 61.78 & 61.75 & 61.66 & 61.84 & 62.52 & 62.74 & 66.35 & \textbf{61.64} \\
Click Bell & 60.15 & 60.37 & 60.32 & 60.19 & 60.18 & 60.07 & 60.00 & \textbf{59.88} & 60.34 \\
Dump Bin Bigbin & 196.62 & 191.63 & \textbf{180.48} & 188.88 & 195.14 & 184.06 & 186.37 & 232.66 & 182.93 \\
Grab Roller & 82.66 & 83.55 & 84.11 & 83.88 & 83.77 & 82.16 & 81.38 & \textbf{80.37} & 83.42 \\
Handover Block & 296.72 & \textbf{280.62} & 280.78 & 289.82 & 293.50 & 299.24 & 373.79 & 444.76 & 281.48 \\
Handover Mic & 206.90 & 214.38 & 213.82 & 205.28 & 210.24 & \textbf{201.29} & 204.35 & 208.13 & 201.65 \\
Hanging Mug & 613.70 & 583.55 & 577.64 & 524.95 & 517.90 & 536.11 & 530.28 & 590.19 & \textbf{506.87} \\
Lift Pot & 102.85 & 102.78 & 102.45 & \textbf{102.37} & 102.76 & 118.38 & 159.67 & 159.82 & 102.39 \\
Move Can Pot & 156.34 & 148.27 & 145.67 & 143.02 & 143.23 & 151.33 & 195.43 & 244.87 & \textbf{142.98} \\
Move Pillbottle Pad & \textbf{135.51} & 138.26 & 138.40 & 138.70 & 141.02 & 139.25 & 155.82 & 177.28 & 141.22 \\
Move Playingcard Away & 105.43 & 102.68 & 102.65 & 102.53 & 102.58 & 102.53 & 105.35 & 105.62 & \textbf{102.47} \\
Move Stapler Pad & 219.05 & \textbf{188.46} & 212.73 & 207.85 & 200.11 & 205.07 & 283.41 & 315.96 & 209.22 \\
Open Laptop & 163.58 & 159.91 & 152.78 & 161.03 & 152.13 & 146.61 & 154.92 & \textbf{139.21} & 163.34 \\
Open Microwave & 598.15 & 568.36 & 558.88 & 545.53 & 696.27 & 782.47 & 939.97 & 976.57 & \textbf{540.67} \\
Pick Diverse Bottles & 155.89 & 141.60 & 145.60 & \textbf{139.80} & 148.47 & 154.40 & 166.20 & 184.84 & 142.51 \\
Pick Dual Bottles & 117.76 & 111.17 & 111.99 & 117.64 & 109.05 & 113.85 & 173.16 & 200.54 & \textbf{108.97} \\
Place A2B Left & 159.04 & 153.40 & 147.43 & 150.68 & \textbf{145.05} & 147.15 & 151.75 & 162.23 & 147.47 \\
Place A2B Right & 159.59 & 153.15 & 160.94 & 153.25 & 157.71 & 157.66 & 153.85 & \textbf{147.28} & 147.30 \\
Place Bread Basket & 228.80 & 235.65 & 224.91 & \textbf{223.08} & 236.15 & 250.52 & 272.64 & 272.39 & 225.35 \\
Place Bread Skillet & 159.94 & 167.38 & 156.36 & 159.64 & 186.86 & \textbf{146.55} & 166.56 & 181.30 & 174.28 \\
Place Burger Fries & 223.08 & 215.15 & 214.81 & \textbf{208.42} & 217.87 & 210.20 & 222.99 & 219.47 & 215.02 \\
Place Can Basket & 367.86 & \textbf{344.61} & 348.20 & 355.27 & 346.96 & 399.32 & 465.18 & 479.58 & 387.95 \\
Place Cans Plasticbox & 233.20 & 221.96 & 222.09 & 233.46 & \textbf{221.51} & 226.72 & 318.39 & 504.95 & 222.19 \\
Place Container Plate & 135.19 & 131.44 & 131.89 & 131.59 & 144.33 & 137.69 & 138.39 & 137.50 & \textbf{128.62} \\
Place Dual Shoes & 253.07 & 252.00 & 262.40 & 244.64 & \textbf{237.30} & 244.91 & 309.33 & 340.09 & 259.79 \\
Place Empty Cup & 155.94 & 157.65 & 157.55 & 153.74 & 154.31 & 153.52 & \textbf{152.41} & 162.92 & 153.16 \\
Place Fan & 160.71 & 165.94 & 148.60 & 154.24 & 150.40 & \textbf{147.58} & 157.98 & 159.72 & 150.40 \\
Place Mouse Pad & 200.15 & 167.61 & 162.94 & 161.48 & 160.19 & \textbf{155.61} & 182.24 & 256.87 & 169.18 \\
Place Object Basket & 302.21 & 279.16 & 280.25 & 293.86 & 281.20 & \textbf{265.26} & 288.01 & 290.49 & 283.14 \\
Place Object Scale & 156.60 & 147.75 & \textbf{145.17} & 147.54 & 147.88 & 152.99 & 164.54 & 170.14 & 151.00 \\
Place Object Stand & 130.46 & 127.71 & \textbf{122.05} & 136.28 & 127.97 & 144.38 & 144.45 & 161.48 & 127.65 \\
Place Phone Stand & 127.70 & 121.75 & 118.73 & \textbf{112.96} & 121.52 & 115.84 & 127.07 & 133.57 & 115.88 \\
Place Shoe & 153.49 & 141.70 & \textbf{141.52} & 142.02 & 144.88 & 142.38 & 145.62 & 149.54 & 145.34 \\
Press Stapler & 133.08 & 112.18 & 106.44 & 107.64 & 103.84 & 100.12 & 96.52 & \textbf{88.76} & 114.33 \\
Put Bottles Dustbin & 698.32 & 677.81 & 642.77 & 666.00 & 651.13 & 695.03 & 677.48 & 892.98 & \textbf{641.40} \\
Put Object Cabinet & 292.33 & \textbf{290.21} & 290.24 & 298.55 & 297.13 & 323.79 & 300.96 & 329.79 & 306.53 \\
Rotate QRcode & 170.36 & 162.11 & 164.88 & \textbf{159.73} & 163.57 & 166.76 & 182.49 & 219.16 & 167.87 \\
Scan Object & 202.25 & 194.85 & \textbf{186.44} & 191.40 & 190.12 & 188.25 & 207.77 & 269.27 & 189.19 \\
Shake Bottle & 92.02 & 92.40 & 92.95 & 92.30 & 91.75 & 91.01 & 96.15 & \textbf{89.09} & 91.34 \\
Shake Bottle Horizontally & 91.89 & 93.33 & 93.29 & 93.90 & 93.34 & 92.23 & \textbf{91.59} & 92.18 & 92.88 \\
Stack Blocks Three & 464.14 & 454.11 & 454.01 & 453.97 & 453.40 & 452.86 & 471.30 & 542.00 & \textbf{446.59} \\
Stack Blocks Two & 286.55 & 285.49 & 285.51 & 284.53 & \textbf{283.58} & 283.97 & 285.03 & 301.37 & 285.14 \\
Stack Bowls Three & 625.10 & 609.91 & 586.92 & 584.71 & 543.19 & 551.92 & 690.94 & 620.42 & \textbf{534.26} \\
Stack Bowls Two & 338.28 & 312.71 & \textbf{298.05} & 308.63 & 311.54 & 319.10 & 312.38 & 312.74 & 305.83 \\
Stamp Seal & 186.50 & 168.20 & 155.73 & 159.19 & \textbf{151.07} & 181.52 & 231.66 & 261.25 & 161.01 \\
Turn Switch & 219.40 & 199.67 & 196.01 & 206.83 & 201.29 & 244.31 & 330.49 & 357.00 & \textbf{186.98} \\
\bottomrule
\end{tabular}
\end{table}

\clearpage

\begin{table}[H]
\centering
\caption{
{\fontfamily{lmr}\selectfont\textbf{Average episode steps on RoboTwin~2.0 in the Randomized setting.}}
Each mean is computed over 100 episodes. $k$ denotes the fixed loop count.
Bold marks the fewest steps in each row.
}
\label{tab:supp_robotwin_steps_random}
\footnotesize\fontfamily{lmr}\selectfont
\setlength{\tabcolsep}{4pt}
\begin{tabular}{@{}l*{9}{r}@{}}
\toprule
Task & $k=1$ & $k=2$ & $k=3$ & $k=4$ & $k=5$ & $k=6$ & $k=7$ & $k=8$ & Adaptive \\
\midrule
Adjust Bottle & 109.69 & 110.19 & 110.21 & 110.72 & 109.38 & 108.07 & 109.75 & \textbf{106.36} & 110.33 \\
Beat Block Hammer & 135.26 & 121.75 & 133.17 & 128.30 & 127.89 & \textbf{120.80} & 162.94 & 206.87 & 133.62 \\
Blocks Ranking RGB & 495.39 & 484.02 & 489.30 & \textbf{480.52} & 484.59 & 481.18 & 495.96 & 512.20 & 489.10 \\
Blocks Ranking Size & 673.62 & 583.52 & 568.61 & 550.08 & 535.14 & 541.44 & 601.56 & 674.67 & \textbf{526.50} \\
Click Alarmclock & \textbf{61.00} & 61.17 & 61.26 & 62.11 & 61.67 & 61.99 & 62.19 & 62.42 & 61.30 \\
Click Bell & 59.98 & 60.17 & 60.17 & 60.11 & 60.06 & 60.00 & 59.93 & \textbf{59.87} & 60.16 \\
Dump Bin Bigbin & 209.66 & \textbf{200.70} & 220.28 & 209.76 & 204.92 & 223.73 & 215.62 & 264.66 & 208.38 \\
Grab Roller & 82.78 & 83.25 & 83.58 & 83.47 & 82.89 & 82.20 & 81.34 & \textbf{81.29} & 83.21 \\
Handover Block & 342.52 & 331.89 & 336.88 & 342.45 & \textbf{309.76} & 350.84 & 395.58 & 524.70 & 331.69 \\
Handover Mic & 201.15 & 202.42 & 201.52 & 205.15 & 201.21 & 201.23 & 204.50 & \textbf{200.73} & 206.17 \\
Hanging Mug & 609.00 & 586.16 & 572.87 & 554.18 & 597.86 & 656.02 & 597.35 & 722.26 & \textbf{548.25} \\
Lift Pot & 107.83 & 105.13 & 104.99 & 104.89 & 105.16 & 111.52 & 149.86 & 175.65 & \textbf{104.62} \\
Move Can Pot & 142.99 & \textbf{142.90} & 145.45 & 145.43 & 148.06 & 161.35 & 210.99 & 270.42 & 148.00 \\
Move Pillbottle Pad & \textbf{137.31} & 137.89 & 137.99 & 138.21 & 138.37 & 141.40 & 146.85 & 155.51 & 137.99 \\
Move Playingcard Away & 105.19 & 102.63 & 102.55 & 102.56 & \textbf{102.44} & 102.46 & 102.97 & 104.36 & 102.49 \\
Move Stapler Pad & 229.21 & 191.94 & \textbf{189.00} & 194.93 & 198.36 & 234.03 & 276.03 & 302.32 & 195.31 \\
Open Laptop & 167.10 & \textbf{143.28} & \textbf{143.28} & 143.39 & 151.87 & 153.26 & 159.28 & 156.91 & 143.81 \\
Open Microwave & 557.27 & 564.76 & 558.76 & 591.24 & 653.53 & 760.46 & 885.37 & 957.92 & \textbf{539.66} \\
Pick Diverse Bottles & 160.61 & 149.76 & 144.31 & 132.40 & \textbf{132.18} & 137.86 & 155.20 & 167.36 & 135.30 \\
Pick Dual Bottles & 126.72 & 114.69 & \textbf{108.69} & 117.72 & 122.87 & 116.25 & 162.47 & 180.14 & 117.32 \\
Place A2B Left & 165.42 & 157.23 & 154.58 & 156.70 & 151.73 & \textbf{151.42} & 159.16 & 164.38 & 151.62 \\
Place A2B Right & 160.39 & 151.73 & 151.97 & 150.91 & 147.27 & 146.12 & \textbf{143.17} & 145.42 & 145.81 \\
Place Bread Basket & \textbf{232.29} & 235.82 & 239.21 & 235.29 & 235.63 & 236.44 & 241.84 & 253.51 & 237.02 \\
Place Bread Skillet & 181.85 & 161.15 & 176.42 & 169.14 & 170.63 & \textbf{158.73} & 170.91 & 178.45 & 170.97 \\
Place Burger Fries & 223.41 & 223.50 & \textbf{217.51} & 220.40 & 221.04 & 222.41 & 231.29 & 231.09 & 226.37 \\
Place Can Basket & 414.81 & 395.04 & 410.70 & \textbf{382.65} & 395.59 & 423.45 & 470.80 & 515.19 & 408.03 \\
Place Cans Plasticbox & 226.49 & \textbf{226.01} & 243.70 & 232.18 & 237.71 & 237.31 & 256.77 & 408.45 & 226.40 \\
Place Container Plate & 140.47 & 133.10 & 136.34 & \textbf{130.94} & 136.64 & 135.05 & 144.74 & 145.21 & 133.57 \\
Place Dual Shoes & 282.33 & 257.41 & 272.29 & \textbf{253.03} & 267.25 & 261.28 & 328.38 & 353.44 & 257.27 \\
Place Empty Cup & 149.75 & 150.72 & 147.15 & 147.93 & 147.86 & 147.90 & 147.39 & 152.34 & \textbf{146.69} \\
Place Fan & 154.64 & 144.06 & 147.22 & 142.69 & \textbf{137.57} & 158.38 & 152.88 & 166.22 & 148.05 \\
Place Mouse Pad & 183.92 & 174.78 & 167.84 & 169.92 & 165.25 & \textbf{160.68} & 179.93 & 244.57 & 173.91 \\
Place Object Basket & 312.11 & 335.02 & 325.05 & 319.37 & 317.04 & \textbf{297.92} & 321.72 & 304.18 & 324.73 \\
Place Object Scale & 138.44 & \textbf{134.88} & 141.20 & 142.87 & 137.55 & 137.24 & 135.65 & 153.26 & 137.56 \\
Place Object Stand & 132.20 & 133.95 & 134.10 & 131.56 & 131.45 & \textbf{131.32} & 139.61 & 167.47 & 142.37 \\
Place Phone Stand & 129.43 & 120.56 & 120.41 & 117.48 & \textbf{117.47} & 126.19 & 126.55 & 144.14 & 123.40 \\
Place Shoe & 150.26 & 139.69 & 139.64 & 142.79 & 142.86 & 142.72 & 157.05 & 158.15 & \textbf{139.62} \\
Press Stapler & 124.95 & 115.60 & 107.95 & 105.43 & 104.22 & 106.27 & 105.34 & \textbf{99.20} & 112.76 \\
Put Bottles Dustbin & 693.82 & 683.59 & 647.14 & 671.47 & 633.08 & 677.04 & 723.54 & 994.56 & \textbf{627.37} \\
Put Object Cabinet & 316.58 & 308.68 & 291.33 & 309.03 & 308.64 & 294.46 & \textbf{288.84} & 330.33 & 306.09 \\
Rotate QRcode & 189.51 & 189.61 & 180.76 & \textbf{175.68} & 185.21 & 185.81 & 203.53 & 234.44 & 182.99 \\
Scan Object & 202.28 & \textbf{192.76} & 196.60 & 202.46 & 202.42 & 206.53 & 244.07 & 283.10 & 202.79 \\
Shake Bottle & 90.05 & 89.36 & 89.43 & 89.27 & 90.07 & 88.64 & 87.59 & \textbf{87.15} & 89.22 \\
Shake Bottle Horizontally & 90.61 & 91.41 & 91.28 & 91.16 & 91.63 & 90.71 & 96.48 & \textbf{90.29} & 91.17 \\
Stack Blocks Three & 473.70 & 460.77 & 467.77 & 467.74 & 466.46 & 475.96 & 503.72 & 602.77 & \textbf{460.30} \\
Stack Blocks Two & 296.88 & \textbf{285.10} & 287.09 & 286.36 & 285.13 & 290.55 & 287.89 & 301.36 & 285.44 \\
Stack Bowls Three & 665.80 & 578.60 & 563.78 & 564.36 & \textbf{536.58} & 578.13 & 558.56 & 664.75 & 571.43 \\
Stack Bowls Two & 308.85 & \textbf{290.88} & 290.93 & 303.63 & 299.84 & 312.43 & 310.63 & 322.03 & 304.18 \\
Stamp Seal & 179.20 & 155.34 & 151.09 & 151.61 & \textbf{144.88} & 173.03 & 212.32 & 234.61 & 150.74 \\
Turn Switch & 214.84 & 195.83 & 201.50 & \textbf{194.41} & 195.11 & 234.08 & 286.92 & 331.45 & 199.57 \\
\bottomrule
\end{tabular}
\end{table}

\clearpage
\section{Inference Latency}
\label{sec:supp_inference_cost}

Table~\ref{tab:supp_inference_cost} reports latency and resource usage per action chunk. Following Fig.~\ref{fig:efficiency_tradeoff}, we profile all eight configurations on one NVIDIA RTX 3090 (24\,GB) in BF16. For these runs, wall-clock timing covers generating a single 16-step action chunk with 10 denoising steps and scheduler updates. GPU busy time measures time spent executing kernels. Peak memory is the maximum GPU memory usage during profiling. Looped-WAM L4 and Fast-WAM have similar median latencies (492.57 and 492.82\,ms), with peak memory of 6.66 and 12.25\,GiB, respectively.

\begin{table}[H]
\centering
\caption{
{\fontfamily{lmr}\selectfont\textbf{Inference latency and resource usage per action chunk.}} L1--L4 denote one to four fixed loops.
}
\label{tab:supp_inference_cost}
\small
\setlength{\tabcolsep}{4pt}
\renewcommand{\arraystretch}{1.08}
\begin{tabular*}{\textwidth}{@{\extracolsep{\fill}}lrrrr@{}}
\toprule
Model & Wall-clock (ms) & GPU busy (ms) & Peak GPU memory (GiB) & TFLOPs/chunk \\
\midrule
Fast-WAM             &  492.82 &  146.34 & 12.25 &  1.999 \\
LingBot-VA           & 1504.52 &  955.00 & 13.75 & 30.107 \\
Motus                & 1490.87 & 1136.44 & 12.07 & 38.483 \\
GigaWorld-Policy &  574.23 &  154.12 & 12.35 &  1.551 \\
\midrule
Looped-WAM L1        &  251.30 &   80.97 &  6.63 &  1.017 \\
Looped-WAM L2        &  334.58 &  107.18 &  6.64 &  1.344 \\
Looped-WAM L3        &  411.95 &  131.30 &  6.65 &  1.672 \\
Looped-WAM L4        &  492.57 &  144.58 &  6.66 &  1.999 \\
\bottomrule
\end{tabular*}
\end{table}

\end{document}